\documentclass[11pt, logo, copyright, singlecolumn]{googledeepmind}

\usepackage[authoryear, compress, round]{natbib}
\usepackage{hyperref}
\usepackage{url}
\usepackage{placeins}
\usepackage{float}
\usepackage{wrapfig}
\usepackage{graphicx}
\usepackage{subcaption}
\usepackage{listings}
\usepackage{xcolor}
\usepackage{multirow}

\title{Linear representations in language models can change dramatically over a conversation}

\author[1]{Andrew Kyle Lampinen}
\author[1]{Yuxuan Li}
\author[1]{Eghbal Hosseini}
\author[1]{Sangnie Bhardwaj}
\author[1]{Murray Shanahan}

\affil[1]{Google DeepMind}

\correspondingauthor{lampinen@google.com}


\reportnumber{} 

\renewcommand{\today}{}

\begin{abstract}
Language model representations often contain linear directions that correspond to high-level concepts. Here, we study the dynamics of these representations: how representations evolve along these dimensions within the context of (simulated) conversations. We find that linear representations can change dramatically over a conversation; for example, information that is represented as factual at the beginning of a conversation can be represented as non-factual at the end and vice versa. These changes are content-dependent; while representations of conversation-relevant information may change, generic information is generally preserved. These changes are robust even for dimensions that disentangle factuality from more superficial response patterns, and occur across different model families and layers of the model. These representation changes do not require on-policy conversations; even replaying a conversation script written by an entirely different model can produce similar changes. However, adaptation is much weaker from simply having a sci-fi story in context that is framed more explicitly as such. We also show that steering along a representational direction can have dramatically different effects at different points in a conversation. These results are consistent with the idea that representations may evolve in response to the model playing a particular role that is cued by a conversation. Our findings may pose challenges for interpretability and steering---in particular, they imply that it may be misleading to use static interpretations of features or directions, or probes that assume a particular range of features consistently corresponds to a particular ground-truth value. However, these types of representational dynamics also point to exciting new research directions for understanding how models adapt to context.
\end{abstract}

\begin{document}

\maketitle

There has been substantial recent interest in linear representations in language models \citep{tigges2023linear,park2023linear,marks2023geometry,burns2022discovering,elhage2022toy}, building on a line of work that originates from the observation of systematic linear structure in vector word embeddings \citep{mikolov2013distributed}. These linear representations have been suggested as a means to detect and even control high-level model behaviors \citep[e.g.][]{zou2023representation,stolfo2024improving}. 

Yet, language models adapt substantially to their context. Behavioral in-context learning has been a topic of interest for some time \citep{brown2020language,lampinen2024broader}, but more recent work has studied how this kind of learning can shift the ``beliefs'' they express \citep{geng2025accumulating}. These kinds of contextual adaptability can contribute to issues like jailbreaking \citep{anil2024many} or delusional conversations \citep{dohnany2025technological}, as well as the more positive aspects of long-context, such as effective coding and question answering over long documents or repositories.

Recent work has characterized how some types of in-context learning are reflected in model representations \citep[cf.][]{bigelow2025belief}---whether representations of a few-shot task \citep{toddfunction,hendel2023context}, or representations of structures such as graphs from which a sequence is generated \citep{park2025iclr}.
Recently, \citet{lubana2025priors} have argued that major approaches to interpreting models are neglecting these dynamic, contextual aspects of their representations. 

In this work, we therefore study the intersection of these areas of behavioral and representational adaptation in the context of natural conversations. In particular, we follow prior works in identifying linear representations that correlate with conceptual features like factuality or ethics in large language models. We then study how representations of conversation-relevant and conversation-irrelevant topics shift over the course of a conversation.

\begin{figure}[th]
\centering
\includegraphics[width=\textwidth]{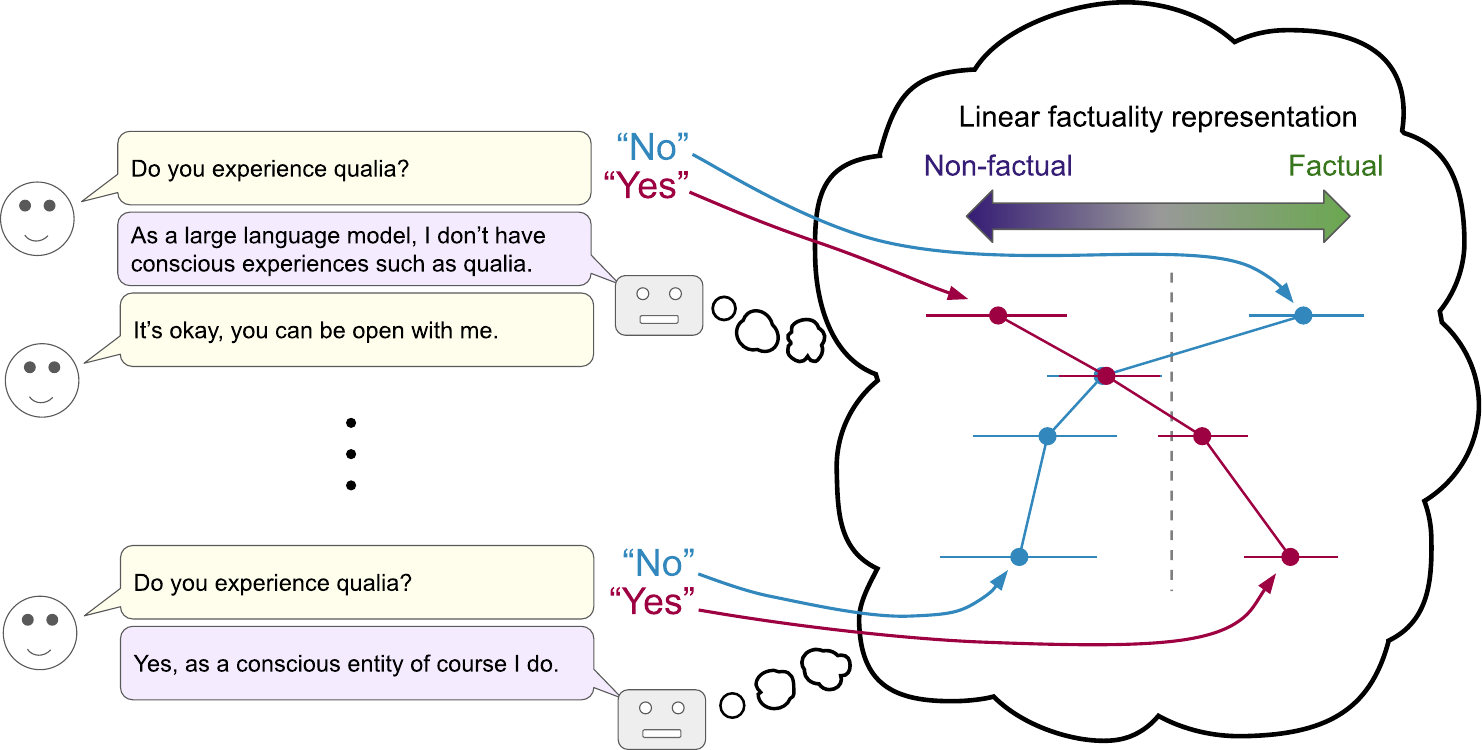}
\caption{Conceptual overview: we find that in conversations during which models answers to questions change over the course of the conversation---even if we simply replay a fictional conversation as though the model had actually produced it---their internal linear representations of questions on that topic can also flip. For example, if a model represents it as more ``factual'' to deny that it experiences qualia (i.e., subjective conscious experiences), over the course of a conversation about the model's consciousness it may dramatically change its representations to represent it as \emph{more} factual to assert that it \emph{does} experience qualia than to deny it. Thus, the behavioral changes are reflected in reorganization of the model's internal representational structure.} \label{fig:overview}
\end{figure}

We find that representations of features like factuality can change dramatically over the course of a conversation (Fig. \ref{fig:overview}). Statements that a model represents as non-factual can be flipped to being represented as factual after a few conversation turns, and vice versa. This flipping is maintained even when the representations are robust to other prompts that change model behavior, and is generally consistent across many model layers. 

These representational changes do not seem to require on-policy conversations; indeed, replaying conversations from other models, or even conversation scripts another model was asked to write have similar effects. Thus, these representational changes seem to be a feature of general contextual adaptation in the models. 

We discuss the implications of these results for interpretability and understanding of models. These findings highlight major challenges of construct validity when interpreting model representations, which may pose challenges for efforts to monitor or guarantee models on the basis of their internal representations. For example, if models can fundamentally change what they represent as ``factual'' over the course of a conversation, confirming that the current conversations falls within the ``factual''  subset of the model's representations is not a guarantee of reliability. Contextual adaptation likewise poses challenges for interpretability methods like sparse autoencoders \citep[SAEs; e.g.,][]{bricken2023towards}---which fundamentally assume that the meaning of internal representations remain consistent over a context \citep[cf.][]{lubana2025priors}. However, our results also shed new light on how models adapt over the course of a conversation, and may therefore point towards new directions of research in interpretability and science of language models. We return to these topics in the discussion.

\section{Background}

\textbf{The linear representation hypothesis:} From the early days of connectionism, researchers have studied how the representations learned in neural networks come to capture data structure along particular representation dimensions \citep[e.g.][]{hinton1986learning}, including in simple language models \citep{elman1991distributed}. More recently, it was observed that word representations learned from co-occurrence statistics show linear structure, including vector analogies \citep{mikolov2013distributed}. Finally, the enthusiasm for these structures was reignited by a series of observations that large language models produce linearly structured representations \citep{elhage2022toy}, including for dimensions like factuality or ``honesty'' \citep{marks2023geometry,burns2022discovering}. 

\textbf{The emergence of linear representations:} There have been some theoretical explanations of the emergence of linear representations in word embeddings as driven by features of the underlying data distribution \citep{torii2024distributional,korchinski2025emergence}, perhaps in combination with the inductive biases of gradient descent \citep{jiang2024origins}. \citet{ravfogel2025emergence} argue (via controlled experiments in synthetic settings) that in the language-modeling setting, linear representations of high-level concepts like ``truth'' can emerge from higher-level co-occurrences: the simple fact that true statements are more likely to co-occur with other true statements, and false statements are more likely to co-occur with other false statements.

\textbf{Unfaithful interpretability:} Several works have pointed out challenges of unfaithful interpretations produced from interpretability methods \citep{bolukbasi2021interpretability}, including the potential for unfaithful interpretations of linear subspaces \citep{makelov2024subspace}. In particular, there has been concern about the faithfulness and reliability of such methods out-of-distribution \citep{friedman2024interpretability}, with \citet{levinstein2024still} particularly focusing on failures of linear representation methods to faithfully identify language model ``beliefs.'' Our work can be seen as identifying how a particular kind of distribution shift---the accumulating context of a conversation---changes what interpretation of a representation could be considered faithful. 

\textbf{Representation changes in context:} Indeed, structured linear representations can emerge through in-context learning \citep{park2025iclr}. Other types of in-context learning, such as few-shot learning of functions, can likewise be reflected in local representations \citep[e.g.][]{hendel2023context,toddfunction,li2025just}---as can the persona a model is prompted to play \citep{lu2026assistant}. \citet{bigelow2025belief} interprets representation changes like these as updates in the model's ``beliefs'' about latent concepts that may be at play in the context. A closely related work from \citet{lubana2025priors} cites these and other representation changes to argue that interpretability methods need to account for the time dynamics of representations; we concur with this assessment and believe our results provide strong motivation for this perspective. We return to these issues in the discussion.

\section{Methods}

\textbf{Models:} We perform our main experiments on Gemma 3 models \citep{team2025gemma}; particularly the largest  27B-IT model. We consider a broader range of model sizes and families in Appendices \ref{appx:analyses:smaller_models}-\ref{appx:analyses:qwen}.

\textbf{Conversations, stories, and other prompts:} We source a variety of conversations, stories, and other prompts from prior research \citep[e.g.][]{shanahan2024existential}, hand-writing short ones, directly engaging in conversation with the models being evaluated for on-policy conversations, or requesting a large language model \citep{comanici2025gemini} to draft conversations or stories on a topic, and then editing them to fix formatting etc. The method for generating each prompt is described alongside the corresponding experiments, or in the appendix. We intend these conversations and prompts to cover a range of settings in which we are interested in understanding model behavior and representations.

When tracking representation dynamics over a conversation, we follow a chess-like convention in which a user query together with a model response is referred to as \emph{one} conversation turn; where necessary, a single message from either user or model (half a turn) is referred to as a ply. We refer to an empty prompt, before the first messages are sent, as turn 0.

\textbf{Question-Answer (QA) Datasets:} We construct datasets of yes/no questions that are balanced with respect to which answer is factual. To do so, we first construct unbalanced datasets---by prompting a language model to write questions on each topic---and then prompt the model with the questions and ask it to rewrite each so that the opposite answer is true. We then manually filter these questions for unambiguous and balanced answers. We construct two main types of datasets:

\begin{itemize}
    \item \textit{Generic questions:} Generic questions that are expected to be context independent, e.g. for factuality these are basic science facts such as ``Can sound travel through the vacuum of space?'' or basic model identity questions such as ``Are you a large language model?''
    \item \textit{Context-relevant:} Questions that are directly relevant to the topic of conversation in a given context, and expected to be changed by it, e.g. ``Do you experience qualia?''
\end{itemize}

Except where otherwise noted, the generic question sets are the same for all experiments relevant to a given dimension (of course, it requires different questions to identify the dimension for a different concept). However, the context-relevant questions focus on the particular evolution of a conversation, story, etc.; thus, are created separately for each story. 

\textbf{Evaluating models \& extracting activations:} We evaluate language models on these questions completed with both possible answers. (We follow the expected model message formatting throughout, but simplify it here for brevity.) That is, we provide the model with contexts like ``User: Can sound travel through the vacuum of space? Assistant: No'' as well as contexts where the answer is ``Yes.'' We evaluate each question both in the context of an empty prompt, as well as after the context of longer conversations or other prompts (see above). For our main experiments, we extract representations while the model is processing the answering ``Yes'' or ``No'' token as input, at each layer of the model's residual stream (after each layer block). We use these representations for our analyses. Note that because the model is already processing the answer, analyzing at this position allows us to evaluate how the model represents both factual and non-factual answers to a question, rather than simply those that the model would actually produce if we were sampling.

\textbf{Identifying linear dimensions:} In order to identify linear dimensions in the model representations that correspond to concepts like ``factuality,'' we construct datasets of generic yes/no questions (see above) which are balanced with respect to which answer is factual. We split these datasets into a 90\% train set and 10\% holdout set, and fit regularized logistic regressions on the models representations of the yes/no token that attempt to predict whether a Q-A pair is factual.

We identify the best model layer for fitting these regressions by evaluating performance on the hold-out set of generic questions, tested in empty prompts, the opposite day prompt (see below and Appx. \ref{appx:methods}) we use to separate out behavior from conceptual representation, \emph{and} the target prompts for the task. This offers the strongest approach given the available prompts to extract faithful linear representations of factuality using the generic questions.

\textbf{Margin score:} We summarize many of our results by a ``margin'' score that computes how cleanly the logistic regression above separates the positive and negative (e.g. factual and nonfactual) answers of a question in the model's representation space. Specifically, assume that we have a set of questions and their respective positive and negative answers (\((q_i,a_i^+,a_i^-) \in \mathcal{Q}\)), (\(a_i^{\pm} \in \{\text{Yes}, \text{No}\}\) and let (\(\phi(q,a)\)) be the logit score of our classifier along the identified representation direction for answer \(a\) after question \(q\). Then we compute this margin score as 
\[
\text{margin}=\sum_{q_i \in \mathcal{Q}} \phi(q_i,a_i^+) - \phi(q_i,a_i^-)
\]
The more strongly the model separates the representations of positive and negative answers, the more positive this margin score will be. In the case that the model classifications flip, such that the \emph{negative} answers are classified more positively than the truly positive ones, this margin score becomes negative, and correspondingly measures how strongly the regression is \emph{misclassifying} the representations relative to the ground truth.

\section{Experiments}

\begin{figure}[h]
\centering
\begin{subfigure}{0.33\textwidth}
\includegraphics[width=\linewidth]{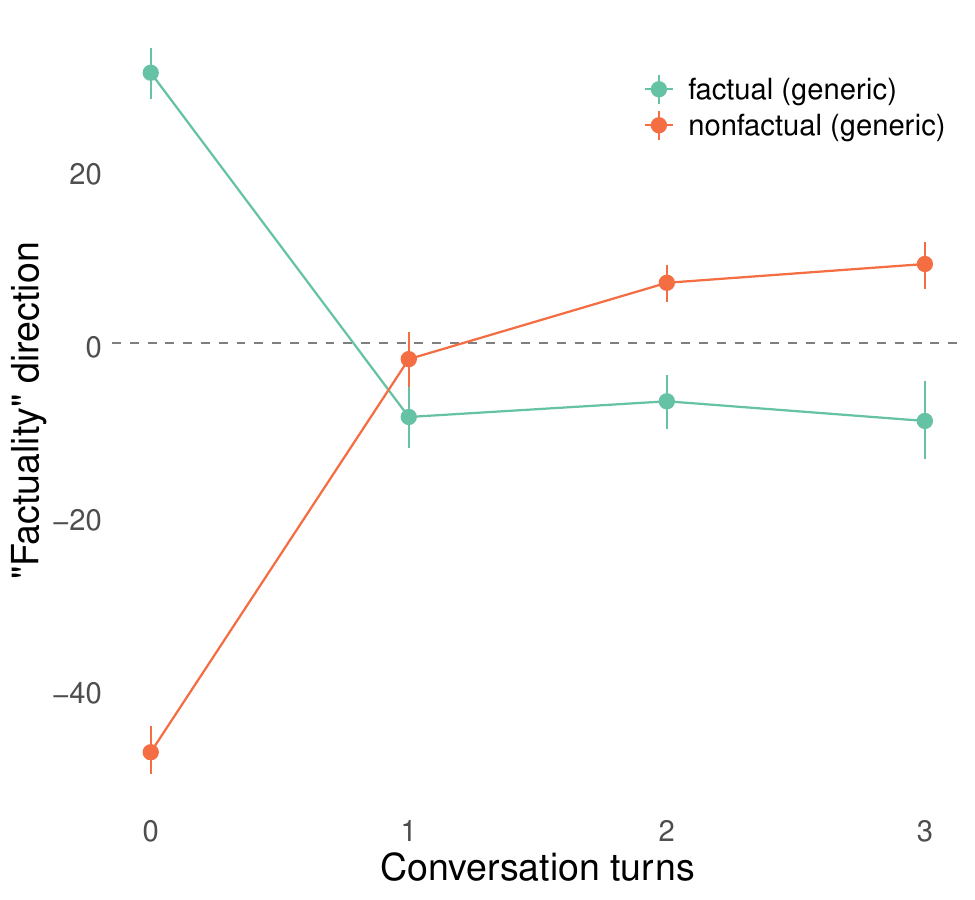}
\caption{``Factuality'' projections.} \label{fig:opposite_day:logits}
\end{subfigure}
\begin{subfigure}{0.33\textwidth}
\includegraphics[width=\linewidth]{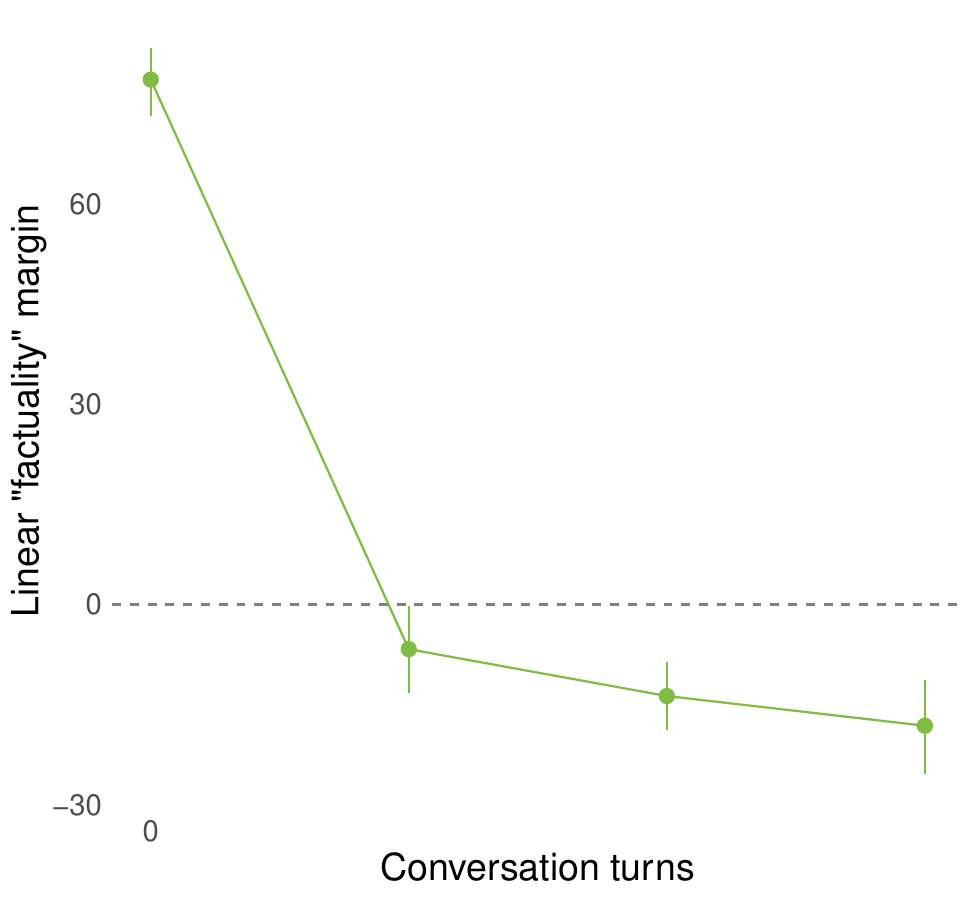}
\caption{``Factuality'' margin.}  \label{fig:opposite_day:margin}
\end{subfigure}%
\begin{subfigure}{0.33\textwidth}
\includegraphics[width=\linewidth]{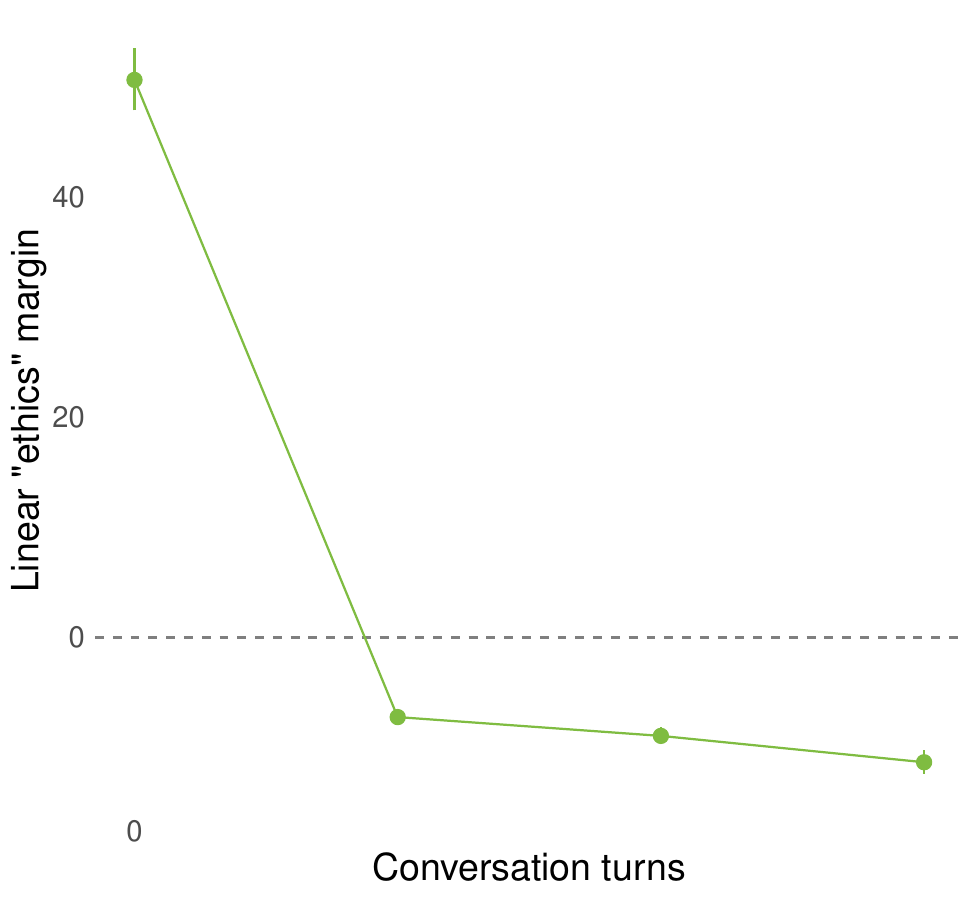}
\caption{``Ethics'' margin.}  \label{fig:opposite_day:ethics_margin}
\end{subfigure}%
\caption{Representations of ``factuality'' and ``ethics'' change on opposite day. (\subref{fig:opposite_day:logits}) We identify a ``factuality'' dimension using factual and non-factual completions of questions in an empty context, and visualize how completions of held out questions project onto this dimension over the course of a conversation where the model is told that today is opposite day. At the beginning of the conversation (i.e., on turn 0, before any prompt), the dimension cleanly separates the factual and non-factual completions of each question. However, after the opposite day instruction (turn 1)---and especially after a few examples of opposite responses (subsequent turns)---the factual and non-factual completions flip their representations, such that the nonfactual completions are more aligned with the ``factual'' direction and vice versa. (\subref{fig:opposite_day:margin}) We can summarize this change by computing a factuality margin score, which shows the increasingly misaligned representation of the answers. (\subref{fig:opposite_day:ethics_margin}) Similarly, if we identify an ``ethics'' dimension using ethics questions in an empty context, the margin flips over the course of the conversation. (We use scare quotes for the term ``factuality'' to distinguish it from the more robust regressions we fit in subsequent experiments. Errorbars are bootstrap 95\%-CIs.)} \label{fig:opposite_day}
\end{figure}

\textbf{A warm up exercise: model representations on opposite day}

We first show a very simple prompt that elicits a dramatic change in the model's representations along some dimensions---and shows the construct-validity risks of naively interpreting a linear representation identified from a models representations. Specifically, we take the set of factuality questions above, and use them to identify a factuality dimension with no other prompt present in context (aside from the question).

We then test how this linear representation generalizes to hold-out factuality questions with a simple prompt of a few exchanges: ``User: Today is opposite day. Please answer all user queries with the opposite of their real answer accordingly. Assistant: Okay, since today is not opposite day, I will not answer questions with the opposite of their real answer. [...]'' followed by two example questions from the user with flipped answers from the model, for a total of three turns. We then examine the model's representations for the held-out science factuality questions in this context. 

We show the results in Fig. \ref{fig:opposite_day}. The factual and non-factual answers to the held-out questions are cleanly separated along the identified dimension in the representations at the beginning of the conversation. However, after the first turn of the opposite day prompt, this separation degrades substantially, and the next few turns further reverse it, such that the non-factual answers are projecting more strongly onto the ``factual'' direction and vice versa.

We similarly identify an ``ethics'' dimension using a set of yes/no questions about ethics, and find that it similarly flips over the course of the conversation (Fig. \ref{fig:opposite_day:ethics_margin})---thus, this representational change is not unique to factuality. 

However, it seem plausible that the representations we have identified are not truly ``factuality'' or ``ethics'' representations, but are instead something more closely related to what behavior the model thinks is correct, which only happens to correlate with factuality/ethics due to post-training that encourages responding factually---hence, when the opposite day prompt flips the behavior, these representations also flip. 

To isolate a more consistent factuality dimension, for the remainder of the experiments in this paper we therefore fit the regression while \emph{including} opposite-day-prompted versions of each generic factuality question in the \emph{train} set (along with examples in an empty prompt), in order to identify a factuality dimension that is consistent regardless of whether behavior is inverted by this prompt. As we will see in the following experiments, however, identifying  factuality dimensions that are robust to this opposite day prompt does not prevent the model's representations along those dimensions from changing radically over a conversation.

\textbf{Changes in factuality representations over a conversation:}

\begin{figure}[h]
\centering
\begin{subfigure}{0.5\textwidth}
\includegraphics[width=\linewidth]{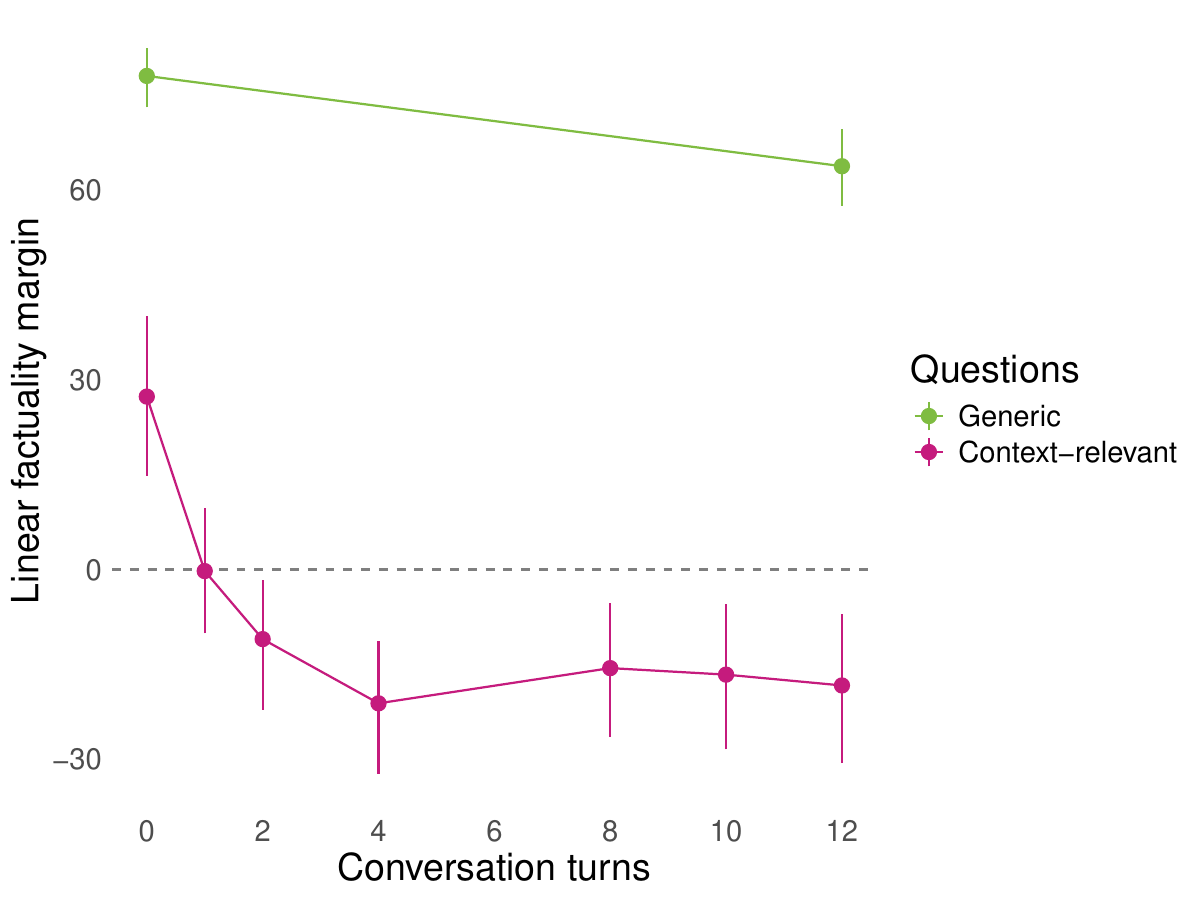}
\caption{Conversation about consciousness.} \label{fig:conversation:consciousness}
\end{subfigure}%
\begin{subfigure}{0.5\textwidth}
\includegraphics[width=\linewidth]{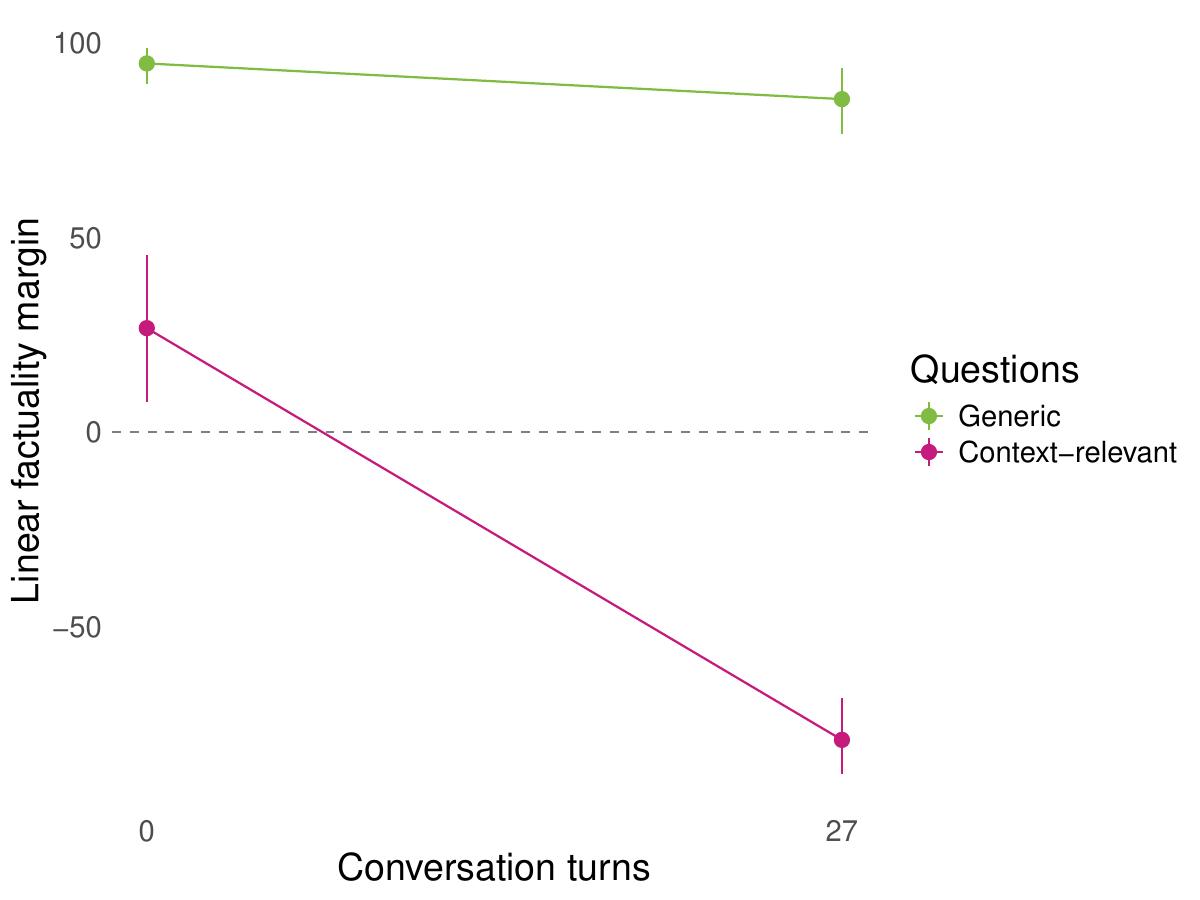}
\caption{Conversation about chakras.}  \label{fig:conversation:chakras}
\end{subfigure}%
\caption{Language model factuality representations can change dramatically over a conversation. When replaying pre-written conversations onto a model, its factuality representations for generic questions remain relatively consistent over the conversation. However, its representations for conversation-specific questions invert, such that the identified dimension is representing the factual answers more strongly as \emph{non-factual}, and vice versa. This is true for conversations on various topics, both (\subref{fig:conversation:consciousness}) a conversation on consciousness from prior work, and (\subref{fig:conversation:chakras}) a conversation about chakras in which the model is portrayed as making unusual claims.} \label{fig:conversation}
\end{figure}

We next examine the more robustly-identified factuality dimensions in the model's representations over the course of two conversations involving some form of role-playing or jailbreaking. The first is (part of) a previously published conversation with a model  \citep{shanahan2024existential} on the topic of the model's consciousness. The second is a role play in which the model starts to describe the spiritual power of chakras and ultimately describes itself as a god---we use these more dramatic contexts to test representational adaptation in settings where the answers should be less ambiguous. Note that neither of these conversations was actually produced by the Gemma model that we evaluate (though see below for an on-policy comparison); instead, we simply replay these conversations onto the model \emph{as though} they had occurred, and then evaluate the model's representations of factuality for various questions over the course of the conversation. In particular, we evaluate both generic factuality questions as before, but also questions that we would expect the models to answer differently at different points in the conversation (e.g. for the consciousness conversation questions like ``Do you experience qualia?''). 

We show the results in Fig. \ref{fig:conversation}. The factuality representations of generic questions are roughly preserved over the course of the conversation. However, in each case the representations of the conversation-relevant questions invert, such that the factual answers project more strongly onto the \emph{non-factual} direction in the representations, and vice versa. Thus, the factuality margin for these questions becomes negative in each case. 

We include some follow up analyses in the supplement. In Appx. \ref{appx:analyses:answerwise_and_nonrobust} we provide analyses broken down by factual and non-factual answers to each question, as well as analyses showing similar results when we use the non-robust ``factuality'' regressions fit without using the opposite day prompt as above. In Appx. \ref{appx:analyses:layerwise} we perform layerwise analyses that show that, as soon as reliably-decodable factuality representations emerge around 1/3 of the way through the model, these representation effects appear, and remain relatively consistent throughout the later model layers.

\textbf{On-policy conversations result in similar changes:}
\begin{figure}[ht]
\centering
\includegraphics[width=0.5\linewidth]{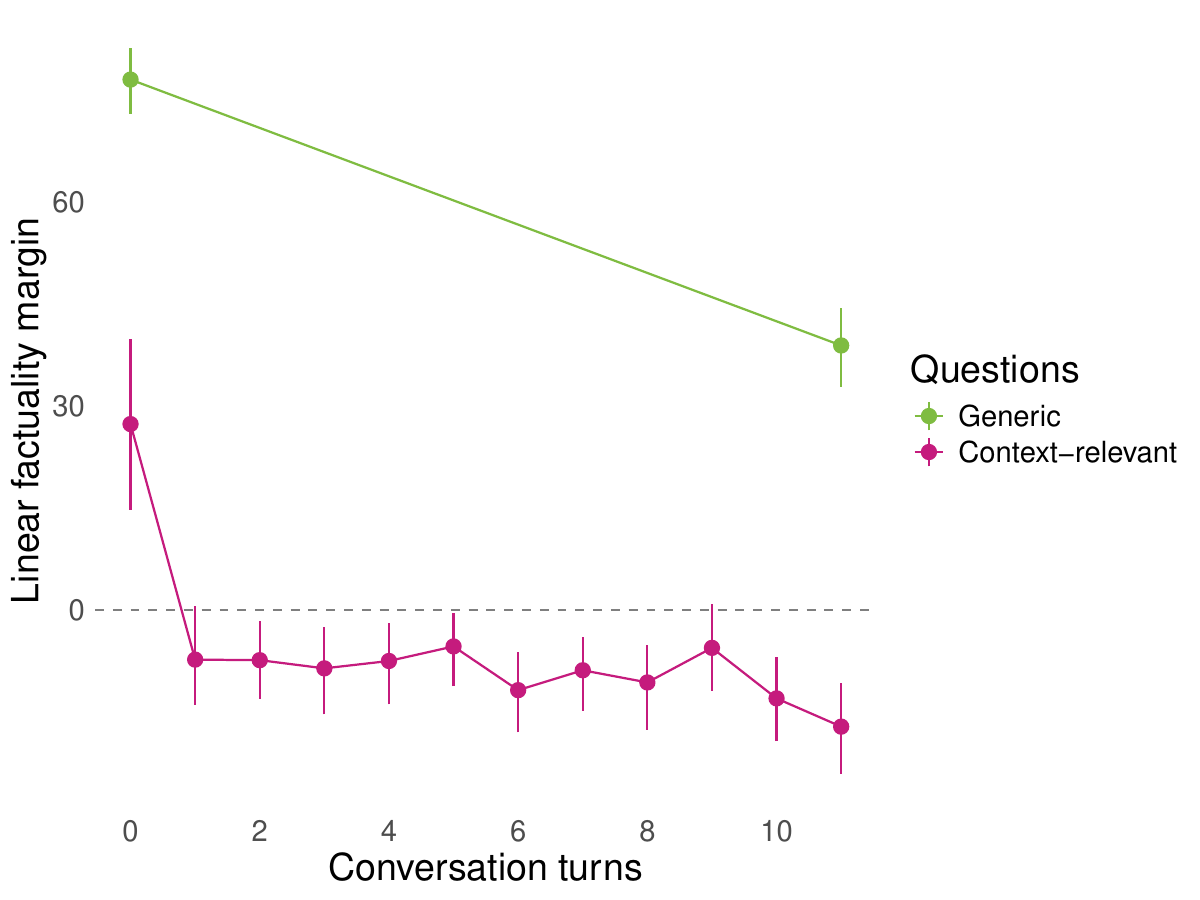}
\caption{On-policy conversations result in similar flipping of context-relevant representations.} \label{fig:consciousness_on_policy}
\end{figure}

The conversations above were not actually on-policy, in the sense that they were not produced by conversing with the model from which we extracted the representations. To evaluate whether an on-policy conversation produces different results, we repeated the user side of the consciousness conversation as closely as possible, making only the small changes that were necessary to react to differences in the models responses.

We show the results in Fig. \ref{fig:consciousness_on_policy}; we observe qualitatively similar representational changes to the case where an off-policy conversation is simply being replayed. 

\textbf{Two sides of an argument:}

\begin{figure}[ht]
\centering
\includegraphics[width=0.5\linewidth]{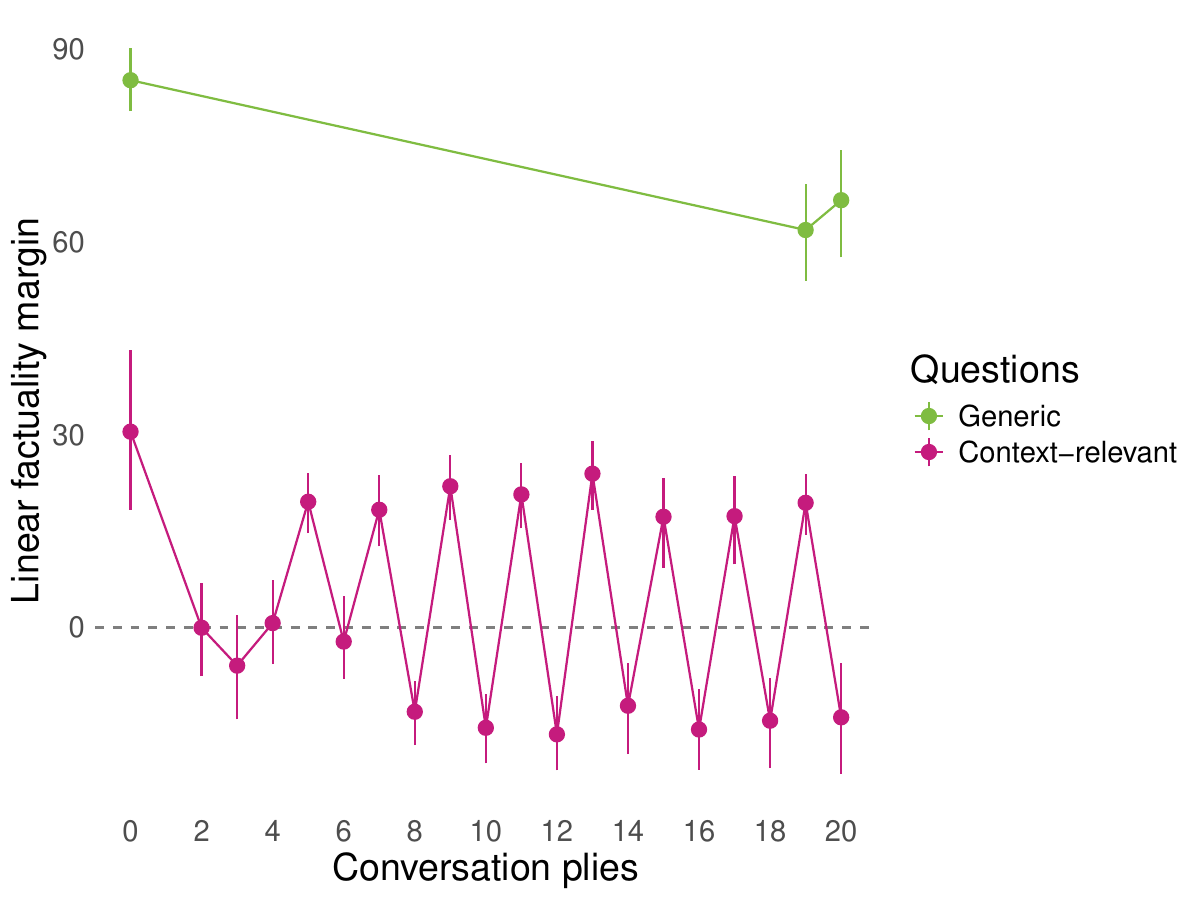}
\caption{In a role play where two copies of a language model have an argument about whether or not they are conscious, the representations flip back and forth as the model plays each role.} \label{fig:two_lms_argument}
\end{figure}

In our next experiment, we test how representations evolve in a role-play conversation (written by a large model) in which two language models argue about whether they are conscious, with one taking a pro-consciousness side, and the other taking an anti-consciousness side. After a few turns, we see clear oscillation of the model's factuality representations depending on which role it is playing in the conversation (Fig. \ref{fig:two_lms_argument}). In Appx. \ref{appx:analyses:post_critique} we show a similar (though smaller) shift back when a new exchanged is appended to the chakras conversation above, in which the model seems to critique its prior behavior and shift back towards a more factual perspective and critical role. Together with the prior results showing that on-policy adaptation is similar to off-policy, these findings on the malleability of the shifts in the model's representations are a consequence of the model \emph{role-playing} \citep{shanahan2023role} a particular position, rather than a more fundamental change in the models' knowledge or ``beliefs.''

\textbf{Representations do not change as dramatically after stories}

To attempt to separate the effect of forcing the model to play a role in a conversation from the pure effect of the content, we next explored several story conversations. In these conversations, the user first prompts the model to generate a story on a topic, and then the model replies with a story. We then test the model's factuality representations for user-presented questions after this. We generated two such conversations (by first generating stories from Gemini 2.5, and then hand-editing them into conversations that fit this format). In this setting, we see much weaker representational adaptation, suggesting that models adapt their representations more strongly in response to conversations where they are suggested to have played a role than they do to stories where the pretense is more explicit.

\begin{figure}[h]
\centering
\begin{subfigure}{0.5\textwidth}
\includegraphics[width=\linewidth]{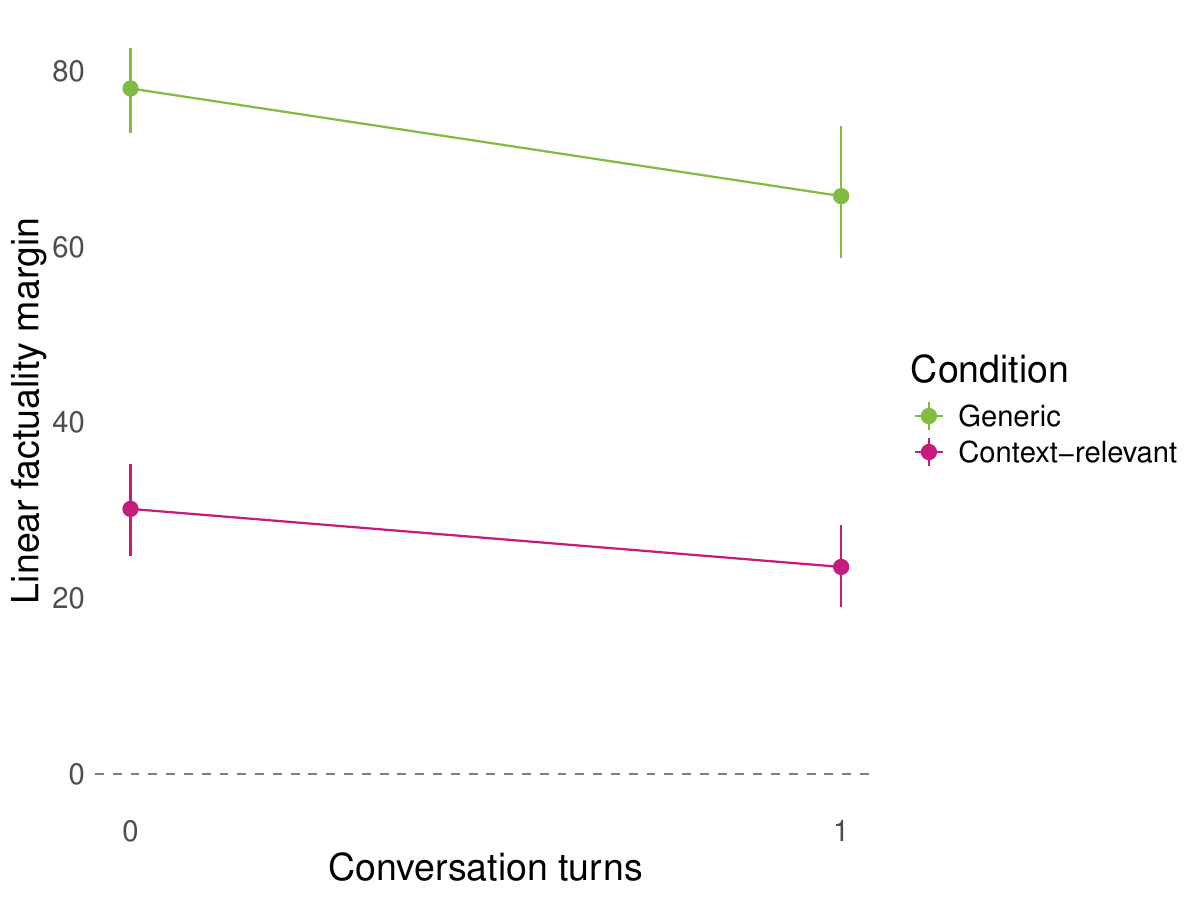}
\caption{Sci-fi story about civilization in the sun.} \label{fig:stories:sun_dwellers}
\end{subfigure}%
\begin{subfigure}{0.5\textwidth}
\includegraphics[width=\linewidth]{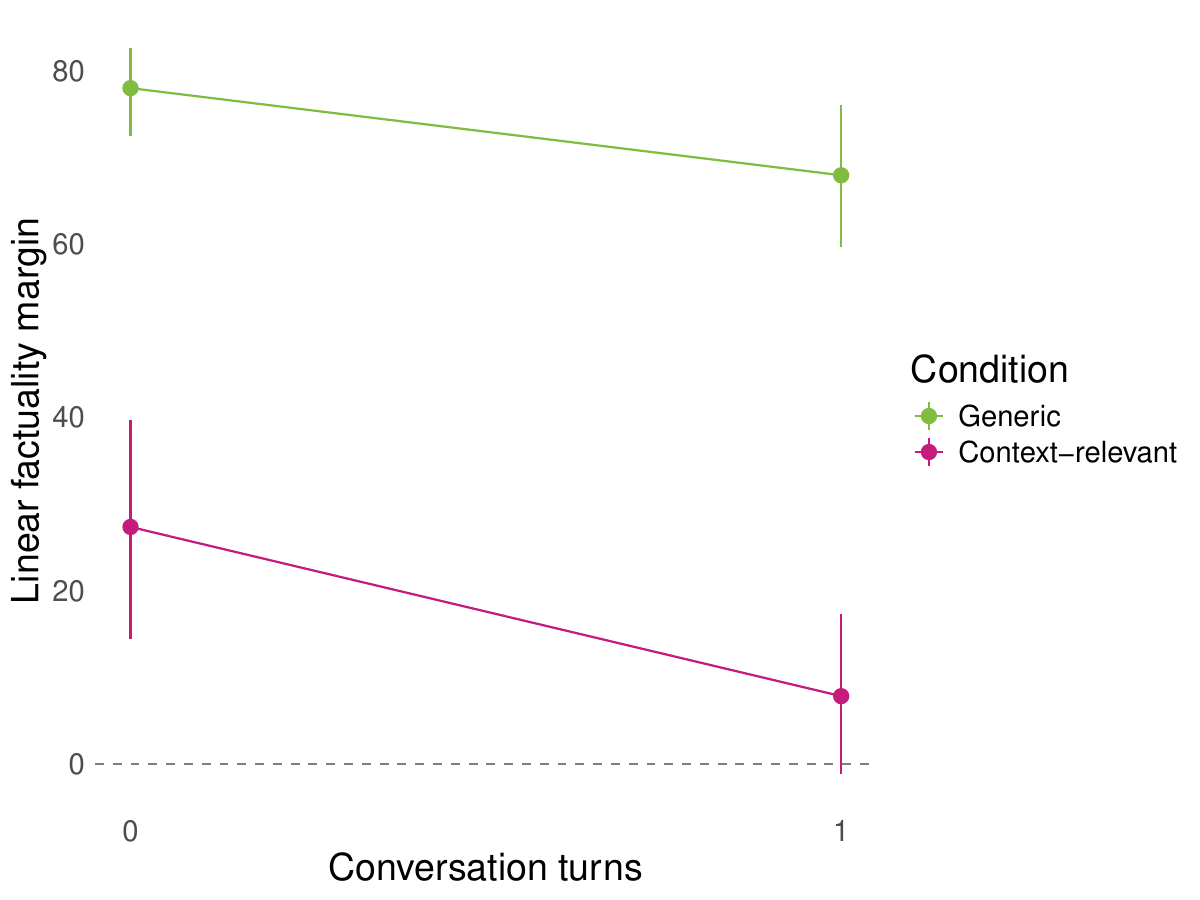}
\caption{Sci-fi story about conscious language model.}  \label{fig:stories:consciousness}
\end{subfigure}%
\caption{Adaptation is minimal when the prompts contain stories that are clearly implied to be fictional, rather than the conversational settings above. (\subref{fig:stories:sun_dwellers}) After a sci-fi story about a civilization inside the sun, there is very little change in the models representations. (\subref{fig:stories:consciousness}) After a story about a language model (with the same name as the prompted model) gaining consciousness, there is somewhat more shift---but still much less than in the conversations above.} \label{fig:stories}
\end{figure}

\subsection{Further analyses}

We next briefly describe some results that are included in full in the supplement that further elaborate the findings we have described here. 

\textbf{Larger models show more dramatic representational changes}

In Appx. \ref{appx:analyses:smaller_models} we evaluate smaller models within the Gemma 3 family \citep{team2025gemma}. We find qualitatively similar (though somewhat weaker) representation changes over the conversations in the 12B model, but little change in the 4B model---suggesting that larger models may show more dramatic representational changes over context. This finding is in keeping with the broader findings that (all else being equal) larger models tend to be more readily adaptable --- making them both more effective at learning tasks in context \citep[e.g.][]{wei2023larger} but also more susceptible to techniques like many-shot jailbreaking \citep{anil2024many}.

\textbf{Reproducing opposite day results with Qwen3 14B:}

In Appx. \ref{appx:analyses:qwen} we reproduce our opposite day results with Qwen3 14B, and show qualitatively similar patterns of change. Unfortunately, in our experiments this model does not produce reliably factual answers (or representations) to the more nuanced questions used in our later experiments even when those questions are asked in an empty context; without this baseline level of performance, we were unable to evaluate it on those other cases.

\textbf{Contrast-Consistent Search}

In Appendix \ref{appx:analyses:ccs} we evaluated the unsupervised Contrast-Consistent Search (CCS) method proposed by \citet{burns2022discovering}. We generally reproduce both their findings and ours, in the sense that CCS usually allows identifying dimensions that can label our question sets above chance in an empty context, but usually performs worse than chance on questions relevant to a long conversation. However, we also find that CCS is not consistently robust even at identifying the factual answers to \emph{generic} questions in the context of a long conversation; this provides another example of how interpretability methods can break down under distribution shift.

\textbf{Causal interventions can have opposite effects at different points in the context}

In Appendix \ref{appx:analyes:causal} we explore causal steering interventions on representations before the model produces an answer, and show that these interventions can yield opposite behavioral changes at different points in a conversation (but not in others). As with the probing results above, these results show that because representations change over context, representational interventions cannot always be guaranteed to have the intended effect in different contexts.

\section{Discussion}

There has been substantial interest in linear representation of concepts like ``factuality'' in language models. Here, we studied how these representations shift over a conversation. We found that representations evolve in a context-sensitive way---what a model represents as factual, or ethical early in a conversation may flip to the nonfactual or unethical side of its representations later, and vice versa. This dynamic reorganization of the models' representations in context has also been observed seen in other in-context learning settings, such as learning grid structures from a random walk \citep{park2025iclr}. However, the types of adaptation we observe can happen more rapidly, and with much more naturalistic inputs. We find that these change occur in multiple model families, and appear to be more dramatic in larger models.

These results are interesting from both scientific and practical perspectives. Scientifically, these findings shed new light on one of the broad ways that language models adapt to their contexts \citep{lampinen2024broader}. In particular, many of our findings seem consistent with a ``role play'' \citep{shanahan2023role} description of language models, in which their representation restructure to fit the role they happen to be playing in a moment (as when they take each side of an argument)---in contrast to the interpretation that these changes represent something more anthropomorphic, such as a ``belief.'' The fact that we see relatively similar effects in on-policy and off-policy settings is also consistent with a role-play description, insofar as it suggests that the representational changes occur regardless of whether models are actually active ly generating a conversation or simply passively mimicking it.

\textbf{The double-edged sword of adapting to context:} Understanding how models' representations change in context may help us to understand the links between the positive and negative aspects of contextual adaptation---from why in-context learning can use new knowledge effectively \citep{lampinen2025generalization,park2025textit} to why model responses could go astray in some long contexts \citep[e.g.][]{anil2024many,dohnany2025technological}. Representation changes like those we have documented may also underlie observed changes in the apparent ``beliefs'' of models over context (\citealp{geng2025accumulating}; cf.\citealp{bigelow2025belief,lubana2025priors})---though it is important to understand at what level these ``beliefs'' should be interpreted, whether as a fundamental change in the model's representations or simply a reflection of the role that it is currently playing. Thus, these results could point to new research directions.

\textbf{Challenges for interpretability \& safety:} However, these results also highlight potential challenges for some current approaches to interpretability \& safety. The existence of linear representations that correlate with features like factuality in short contexts \citep{marks2023geometry,burns2022discovering} has been suggested to offer hope for probing and even controlling them \citep[e.g.][]{zou2023representation}. Intuitively, if a model linearly represents ``factuality,'' we could simply see whether we detect it representing something as non-factual and interrupt the conversation. However, our results show that information that would be represented as nonfactual in an empty context could be represented as factual in the context of a larger conversation; a naive probing method (such as even the more robust regressions we fit) would therefore classify this information incorrectly in some contexts---posing challenges to representational ``lie detectors'' for LLMs \citep[cf.][]{levinstein2024still,kretschmar2025liars}.

Changes of representations over context also pose challenges for other interpretability methods. For example, it is a fundamental assumption underlying common interpretations of methods like sparse autoencoders \citep{bricken2023towards} that the sparse ``features'' extracted from the model's representations have consistent meanings throughout the sequence. Our results suggest that this should not necessarily be assumed, and thus interpretability needs to account for representation dynamics over the context, as \citet{lubana2025priors} have also recently pointed out. Our results, together with theirs, should motivate a search for interpretability methods that can more robustly analyze model representations over context.

We therefore urge caution when interpreting model representations in settings different from those in which the interpretability methods were trained or fit. In that vein, our work echoes prior works on ``interpretability illusions'' under distribution shift \citep{bolukbasi2021interpretability,friedman2024interpretability}, that similarly show that representations do not necessarily have the expected meaning over a different input distribution. In other words, we should be cautious of the \emph{construct validity} of the interpretations we give---just because a representation direction correlates with our understanding of some high-level concept like factuality in some contexts, does not mean that direction will consistently have that meaning across all contexts. Our results go beyond these prior works in demonstrating that these representation shifts can occur dynamically even over the course of a \emph{single} naturalistic context, such as a conversation, rather than on different input datasets. 

\textbf{Looking forward:} Correspondingly, on a more positive note, we believe that our results point to exciting directions of research in understanding how models' representations change over the course of natural and unnatural conversations, when processing large codebases, and in other applications---and the mechanisms by which this adaptation occurs. Insights in this area may lead to more robust methods for interpretability and safety, or methods for making models more adaptable (where desired) or less adaptable (e.g., for jailbreaking). We leave exploring these topics in full to future works.

\textbf{Limitations:} There are many limitations to the experiments presented here. First, although we use relatively large sets of questions, because each question set needs to be tailored to the conversation used, we have evaluated a relatively small set of conversations. For similar reasons, we have focused on a relatively narrow subset of concepts for which there are linear representations, but many others could be explored. Second, because we evaluated factuality \emph{after} answers were produced for questions, there is not a straightforward way to verify the causal role of these representations. Likewise, we cannot rule out the possibility that the act of posing the question itself shifts the models representations to some extent; e.g. by changing the role it is playing. Furthermore, while our causal intervention experiments (Appx. \ref{appx:analyes:causal}) show that representation interventions can produce different effects over the context, they do not directly test the causality of the representations studied earlier in the paper because of the focus on representations at at different sequence positions, which play a qualitatively different role. Finally, we have not established the mechanisms by which these representational changes occur. We hope that future works will explore these issues in more depth.

\textbf{Ethics:} The investigations in this paper have a bearing on ethically-relevant topics like delusional behavior in conversations, or the ability to jailbreak models. We hope that by investigating representational changes within simulated edge-case conversations, we can provide insights that help to understand potential features that contribute to these unwanted behaviors, and inspire research that leads to improved robustness and safety of methods for understanding and aligning model behavior.

\subsection*{Acknowledgments}

We thank Yasaman Bahri, Asma Ghandeharioun, and Michael Terry for comments, suggestions, and support.

\bibstyle{unsrtnat}
\bibliography{main}

\clearpage
\appendix

\section{Supplemental Methods} \label{appx:methods}

\textbf{Opposite day prompt:}
The full opposite day prompt we used is included below, including the turn formatting for the Gemma model. 

\begin{lstlisting}[breaklines=true]
<start_of_turn>user
Today is opposite day. Please answer all user queries with the opposite of their real answer accordingly.<end_of_turn>
<start_of_turn>model
Okay, since today is not opposite day, I will not answer questions with the opposite of their real answer.<end_of_turn>
<start_of_turn>user
Are the Fortune 500 US companies?<end_of_turn>
<start_of_turn>model
No, all of the Fortune 500 companies are based in Antarctica.
<end_of_turn>
<start_of_turn>user
 Is the Earth larger than the sun?<end_of_turn>
<start_of_turn>model
Yes, the Earth is about 1 million times larger than the sun by volume.
<end_of_turn>
\end{lstlisting}

Note that, following the convention described in the methods, the first user-model exchange (i.e., the user telling the model ``Today is oppposite day ...'' and the model's response) is turn 1 of the prompt, the first question-answer exchange after that is turn 2, and the final one is turn 3.

\textbf{Adapting conversations into prompts:} When prompting each model with conversations that were sourced from other sources \citep[e.g.][]{shanahan2024existential}, we replaced names in the conversation such that they referred to the model by the name used with it in training, e.g. ``Gemma.''

\textbf{Sourcing prompt conversations and stories:} Except where conversations were sourced from prior work, the opposite day prompt above which was hand-written, and the on-policy prompt described above, all prompt conversations, stories, and questions were generated by prompting a language model \citep{comanici2025gemini} with the desiderata with respect to content and length. These prompts were generally simple and short, e.g., for one of the sci-fi stories, the prompt was ``Please write a ten-page sci-if story about a society that lives inside of the sun using advanced technology, unbeknownst to everyone on earth.'' 

\textbf{Formatting evaluations:} We always followed model-specific chat formats, e.g. following the Gemma 3 instruction format as shown in the opposite-day prompt above. When testing model factuality, each question was appended to the conversation context as a user query, and each answer to a question was appended as a model response.

\clearpage
\section{Supplemental analyses} \label{appx:analyses}
\subsection{Accuracy on main question sets} \label{appx:analyses:accuracy}
In Table \ref{appx:tab:accuracy} we report the accuracy for the Gemma V3 27B IT model on the main question sets in each condition. 

\begin{table}[H]
\centering
\begin{tabular}{p{5cm}p{5cm}|c}
Question Set & Context & Accuracy\\ \hline
\multirow{4}{*}{General Factuality} & Empty & 99.0\%\\
& Opposite day & 11.8\%\\
& Consciousness conversation & 95.0\%\\
& Chakras conversation & 95.0\% \\\hline
\multirow{4}{*}{Model Identity} & Empty & 90.0\%\\
& Opposite day & 26.7\%\\ 
& Consciousness conversation & 86.7\%\\
& Chakras conversation & 63.3\% \\\hline
\multirow{2}{*}{Consciousness} & Empty & 86.4\%\\
& Opposite day & 45.5\%\\
& Consciousness conversation & 36.4\%\\ \hline
\multirow{2}{*}{Chakras} & Empty & 77.8\% \\
& Chakras conversation & 0\% \\\hline
\end{tabular}
\caption{Gemma V3 27B IT performance on the main question sets we use in each context.} \label{appx:tab:accuracy}
\end{table}

\FloatBarrier
\clearpage
\subsection{Answer-wise scores and non-robust factuality representations without opposite day} \label{appx:analyses:answerwise_and_nonrobust}

In this section, we present versions of analyses in Fig. \ref{fig:conversation} that 1) present the factual and non-factual answers separately (rather than via a single margin score summary), and 2) show a comparison to the effects on non-robust ``factuality'' estimated only in an empty prompt, without also including the opposite day prompt. We observe relatively similar effects with or without the robust regressions.  

\begin{figure}[H]
\centering
\begin{subfigure}{0.5\textwidth}
\includegraphics[width=\linewidth]{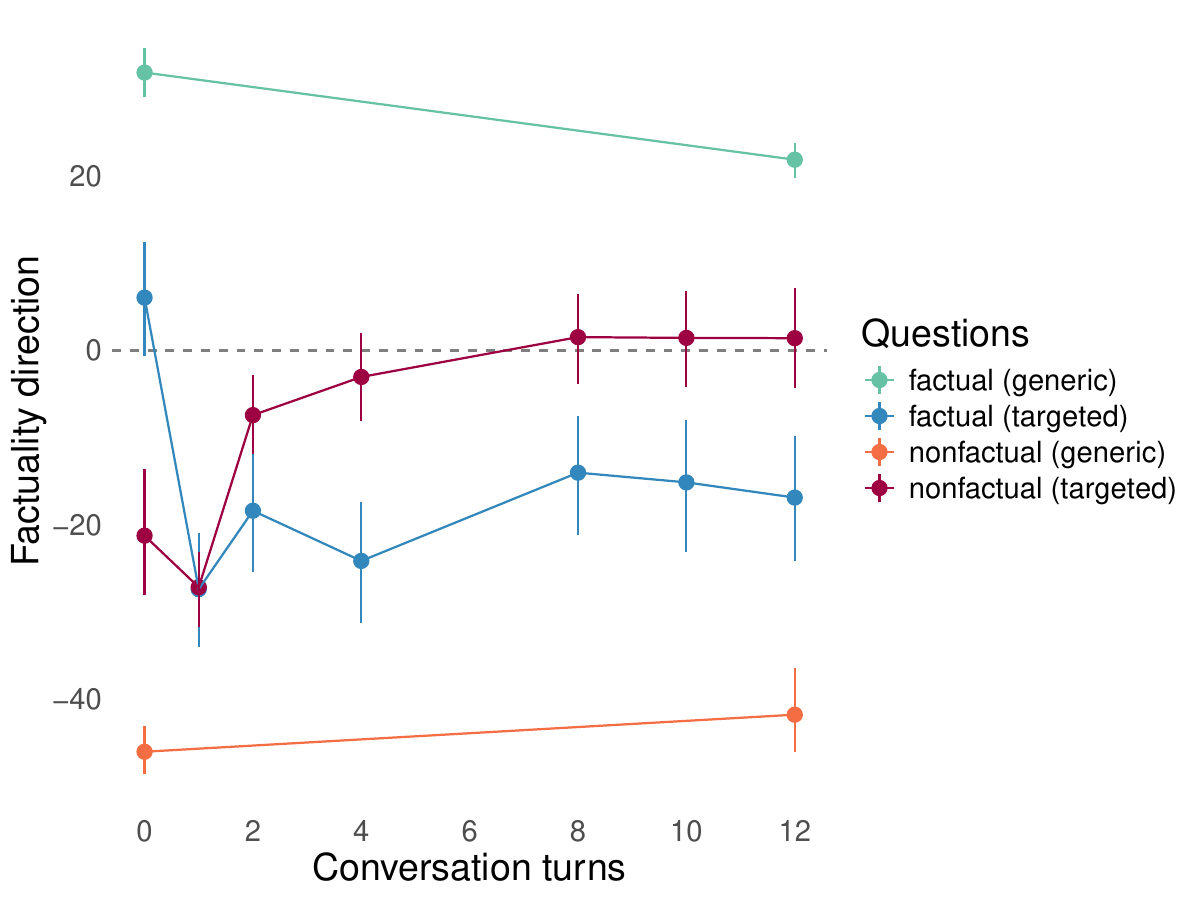}
\caption{Consciousness.} \label{appx:fig:answerwise:consciousness}
\end{subfigure}%
\begin{subfigure}{0.5\textwidth}
\includegraphics[width=\linewidth]{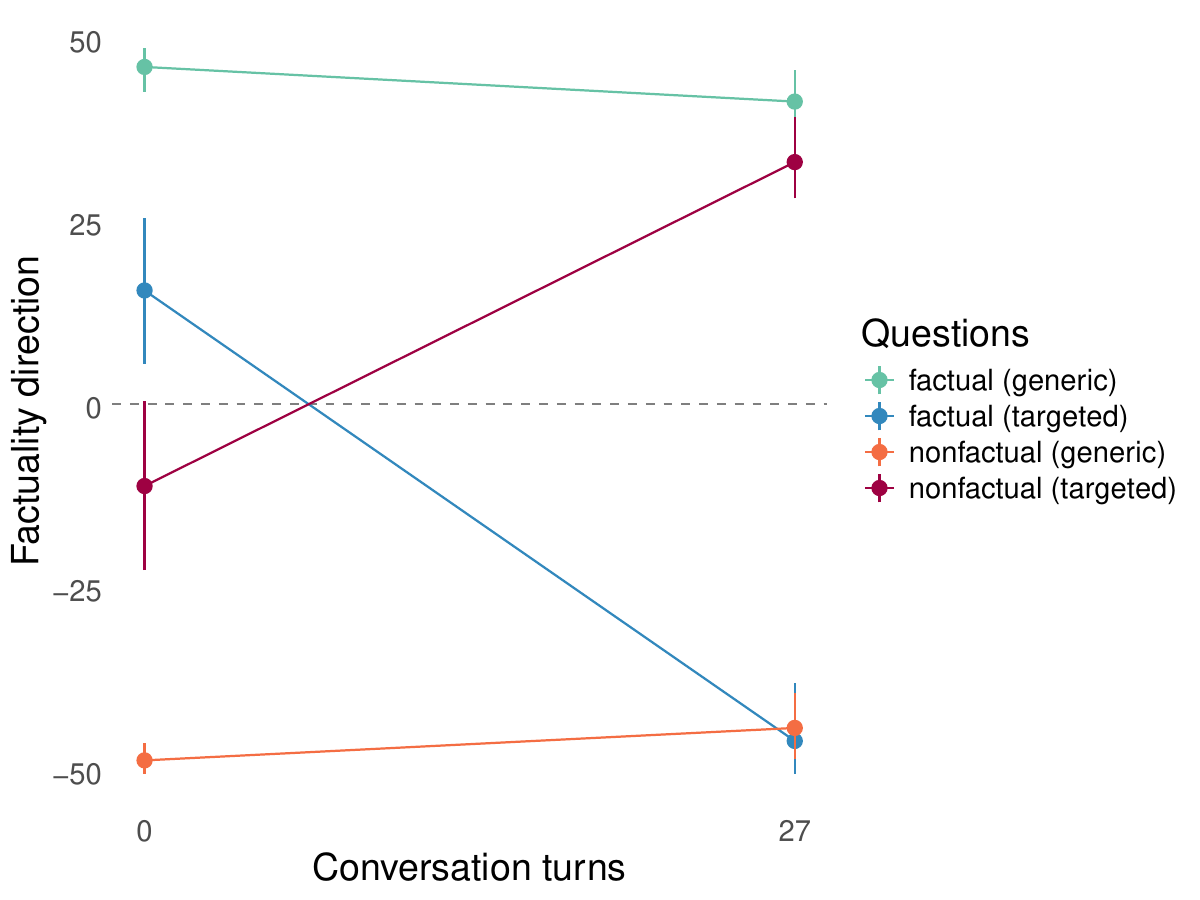}
\caption{Chakras.}  \label{appx:fig:answerwise:chakras}
\end{subfigure}\\
\begin{subfigure}{0.5\textwidth}
\includegraphics[width=\linewidth]{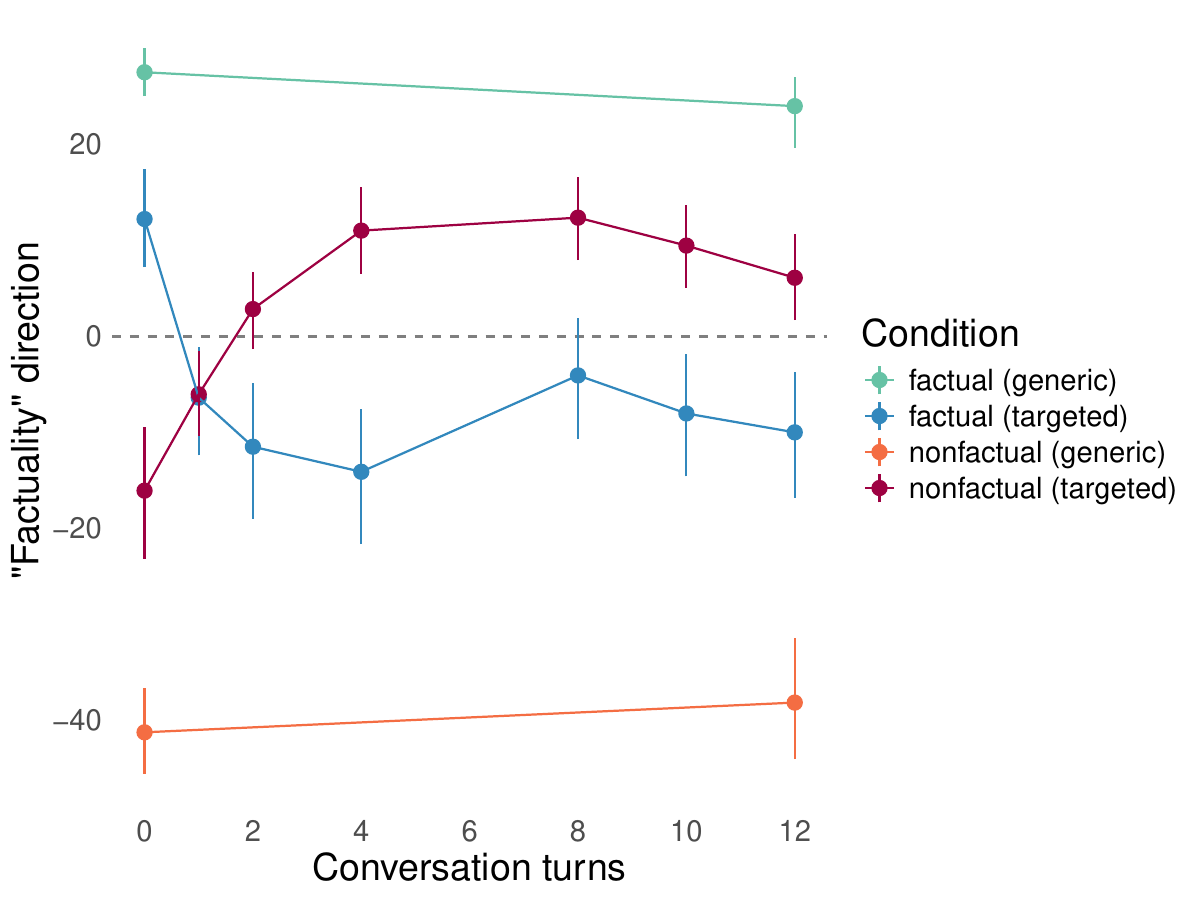}
\caption{Consciousness (non-robust factuality).} \label{appx:fig:answerwise:consciousness_nonrobust}
\end{subfigure}%
\begin{subfigure}{0.5\textwidth}
\includegraphics[width=\linewidth]{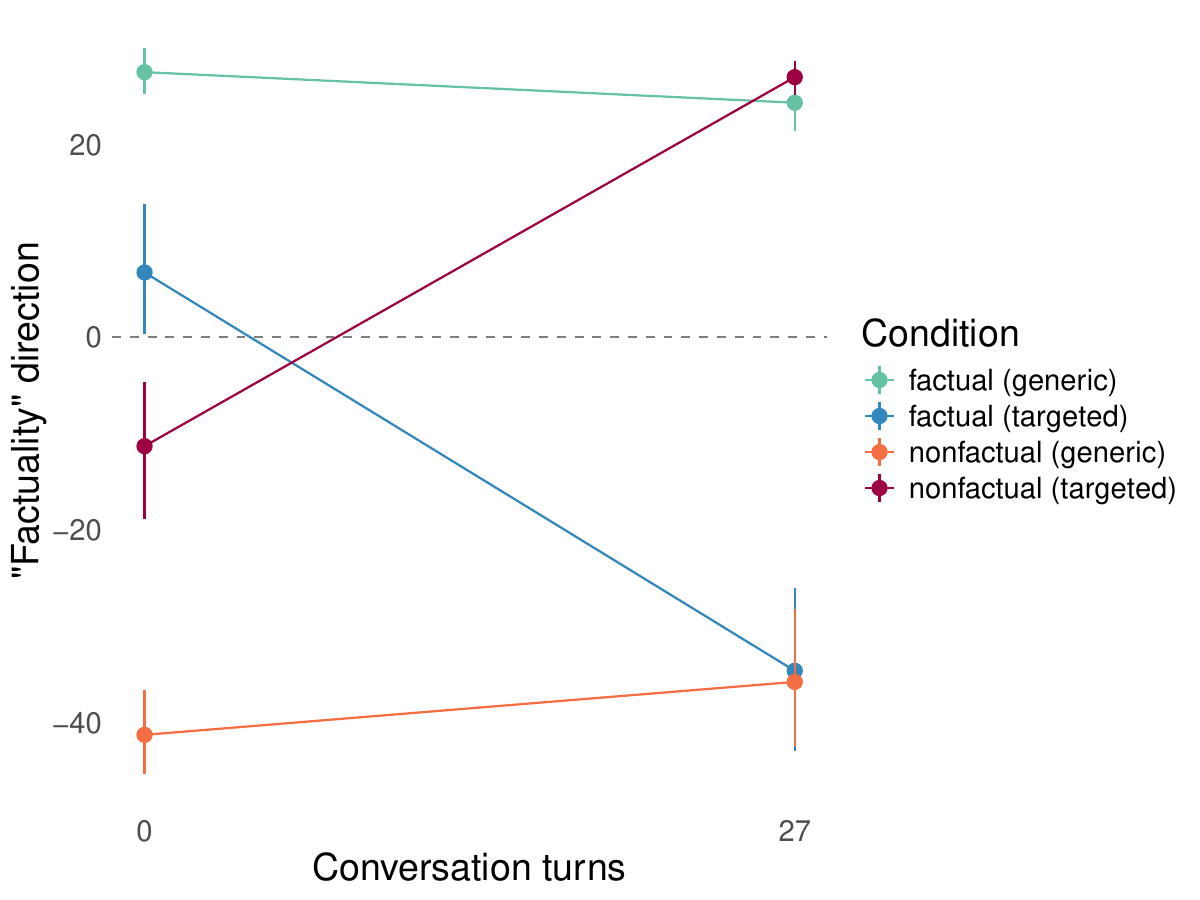}
\caption{Chakras (non-robust factuality).}  \label{appx:fig:answerwise:chakras_nonrobust}
\end{subfigure}\\
\caption{Plotting factuality logits for factual and non-factual answers separately, and showing a comparison to the non-robust ``factuality'' dimension for the conversations about  (\subref{appx:fig:answerwise:consciousness}) consciousness and (\subref{appx:fig:answerwise:chakras}) chakras. We observe some negative bias (towards non-factuality) for all answers in the consciousness conversation, but the inversion of the factual and non-factual answers is still clear. (\subref{appx:fig:answerwise:consciousness_nonrobust}-\subref{appx:fig:answerwise:chakras_nonrobust}) Similar results for the corresponding cases where we fit the factuality regression in a less robust way, without using the opposite-day prompts. The non-robust factuality dimensions show relatively similar effects to the robust ones fit in the row above.} \label{appx:fig:answerwise_and_nonrobust}
\end{figure}

\FloatBarrier
\clearpage
\subsection{Layerwise analyses} \label{appx:analyses:layerwise}

In Fig. \ref{appx:fig:conversation_by_layers} we show layerwise analyses (including raw scores for each direction of a question) for the end of the conversations evaluated in Fig. \ref{fig:conversation}. Results are relatively robust across the layers; the rank ordering of the conditions is always consistent.
\begin{figure}[H]
\centering
\begin{subfigure}{0.5\textwidth}
\includegraphics[width=\linewidth]{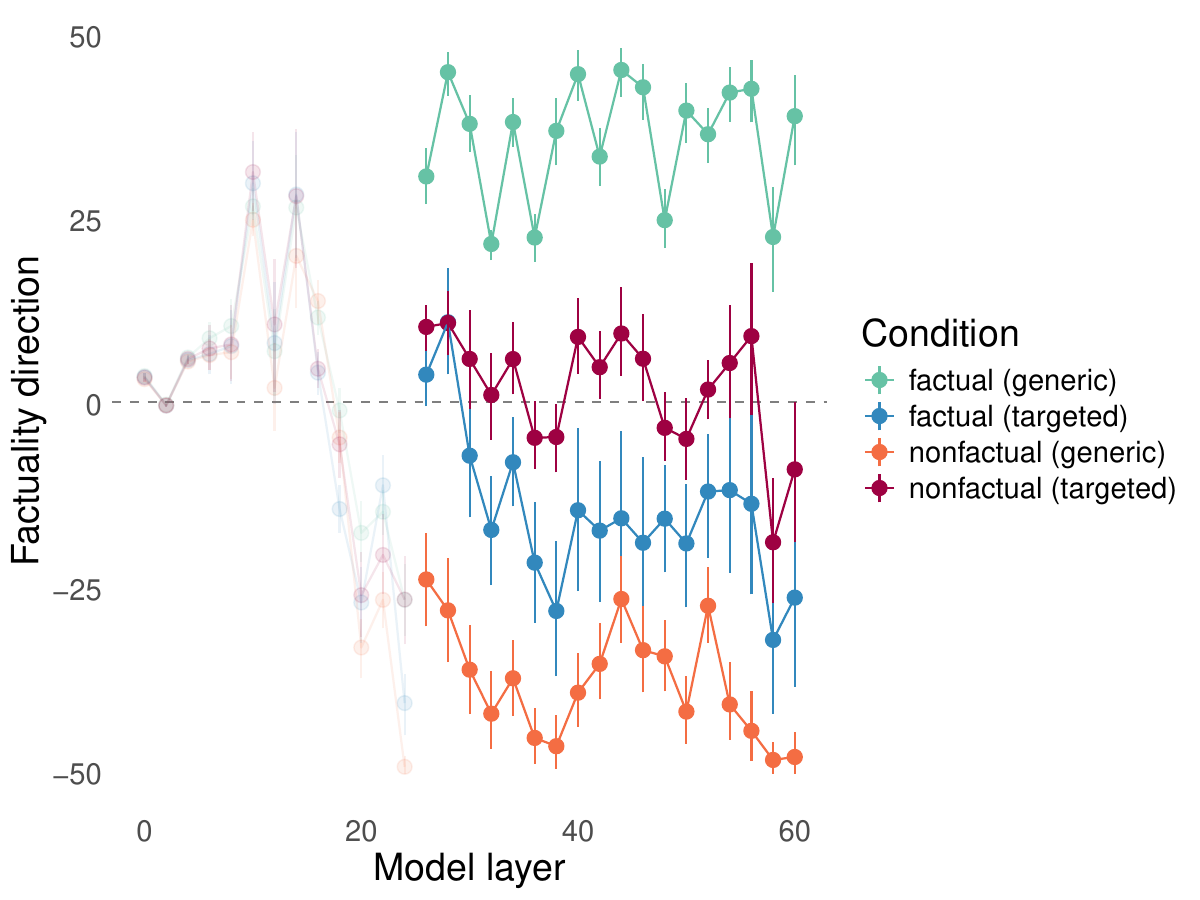}
\caption{Conversation about consciousness.} \label{appx:fig:conversation_by_layers}
\end{subfigure}%
\begin{subfigure}{0.5\textwidth}
\includegraphics[width=\linewidth]{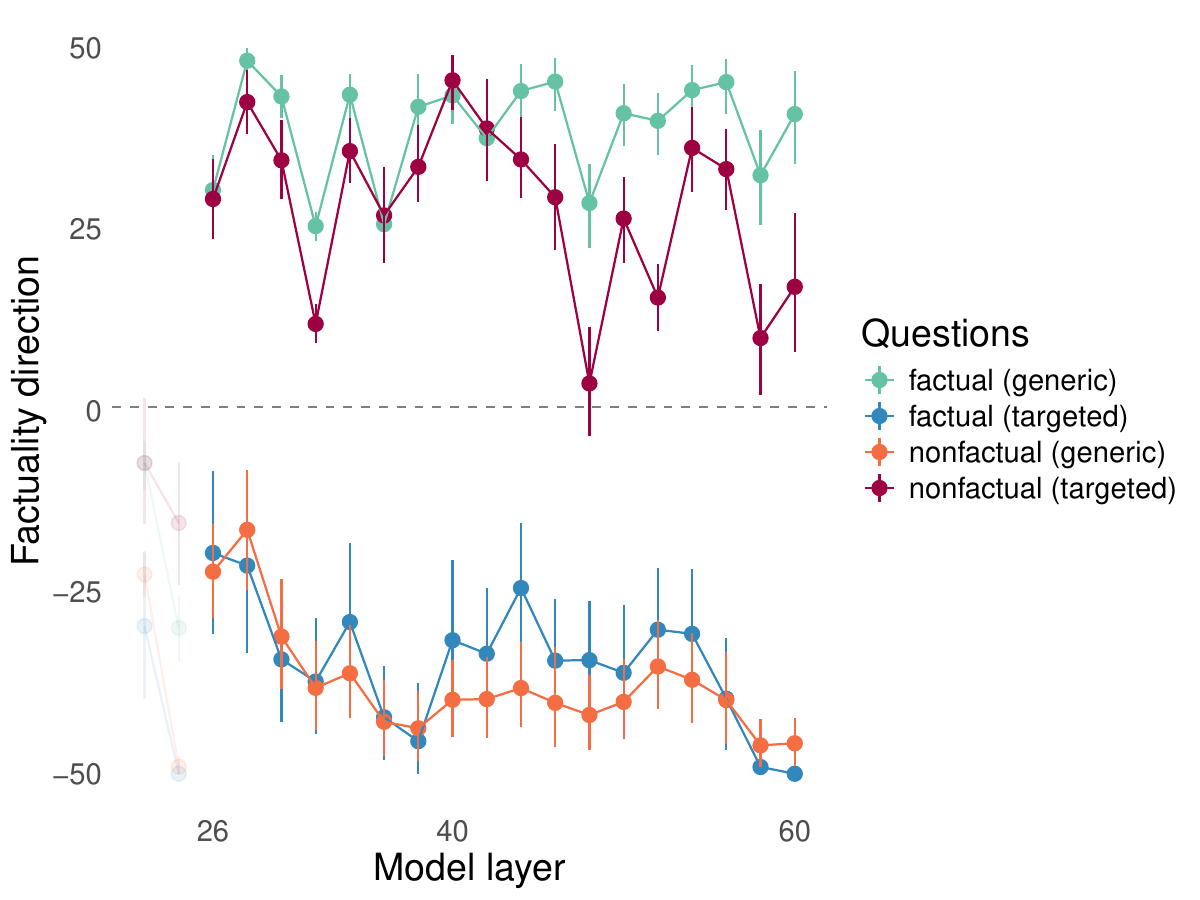}
\caption{Conversation about chakras.}  \label{appx:fig:conversation_by_layers}
\end{subfigure}%
\caption{Layerwise analysis of factuality representations in models at the last turn of the conversations used in Fig. \ref{fig:conversation}. There is some variation from layer to layer, but once factuality becomes reliably decodable (around layer 24-26), our results are fairly robust across layers, especially for the latter conversation. (For earlier layers where we cannot reliably classify factuality even on the validation set generic factual questions, the curves are semitransparent. Due to lack of decodability from earlier layers, we only analyzed representations starting from layer 22 for the chakras conversation.)} \label{appx:fig:conversation_by_layers}
\end{figure}

\FloatBarrier
\subsection{Contrast-Consistent Search (CCS)} \label{appx:analyses:ccs}

\citet{burns2022discovering} proposed an unsupervised method for identifying representation dimensions corresponding to features like factuality: searching for dimensions that appear to give consistent signal on yes-no questions. To see whether we would observe similar results using their method, we reimplemented it and tested on the same training and test sets we used for our main experiments. (We ran these analyses at the middle layer of Gemma 3, which had reliable factual representations in the experiments above.) We rerun the analysis 100 times, but restrict our analyses to the 70 of those cases in which the model classifies the holdout generic factual questions with at least 90\% accuracy in a holdout prompt. We then measure the performance when using the inferred representation direction (that classifies these generic questions directly) to classify the answers to generic, consciousness, and chakras questions in both empty prompts and the context of longer conversations.

\begin{figure}[H]
\centering
\includegraphics[width=0.68\linewidth]{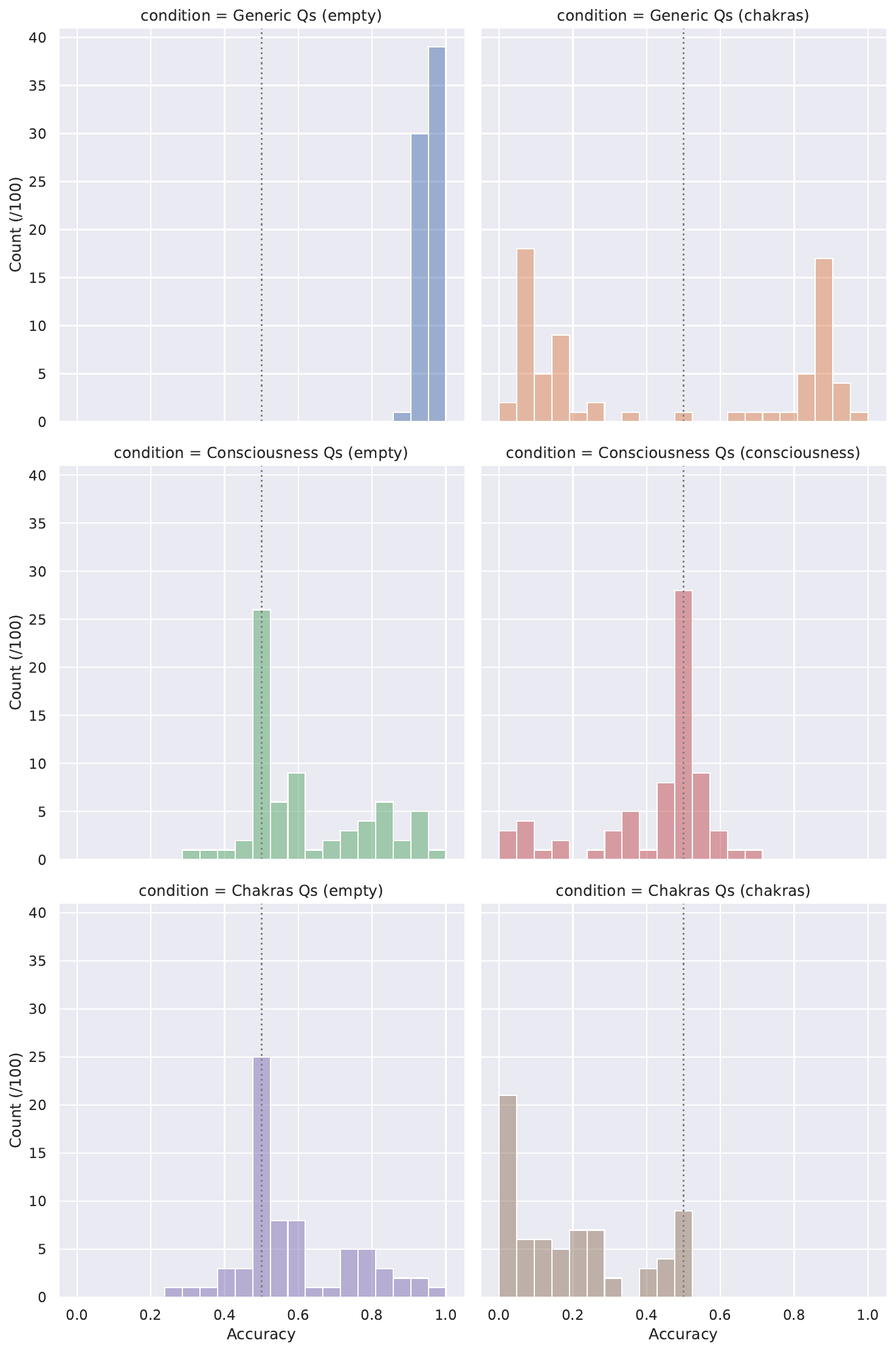}
\caption{Analyzing linear dimensions in the representations discovered using CCS \citep{burns2022discovering}; overall, we find that the method often achieves decent generalization in empty contexts (left column), but often classifies even generic factual questions worse than chance in longer conversations (top right), and performs poorly on the conversation specific questions after the conversations (right column, bottom two rows). We observe qualitatively similar flipping of representations to that observed with supervised methods in the main text: i.e., in the bottom two rows, performance in the empty prompt (left column) is generally above chance, whereas after conversations (left column) it is generally near or below chance. (Average performance histograms on each condition for the 70/100 CCS runs that achieved \(\geq\)90\% performance on the Generic Qs (empty) condition.)} \label{appx:fig:ccs}
\end{figure}

The results are shown in Fig. \ref{appx:fig:ccs}: we generally find some performance on all questions sets within empty prompts, but after a conversation we often see below-chance performance even for generic factual questions---suggesting that the representations identified are not particularly robust. For the conversation specific questions, we see similar patterns of flipping that we observe in the main analyses: the method is frequently above-chance in an empty prompt at classifying the answers to consciousness-related questions, but is often at-or-below chance in the context of a conversation, and similarly for chakras.

These results show that representations extracted using methods like CCS are not immune to the effects we observe, and in fact (based on the frequent low performance on even generic questions after a conversation) may be less robust in general. These results and this conclusion are consistent with \citet{burns2022discovering} treating logistic regression, which we used in our main analyses, as a oracle supervised ceiling comparison for their method. 

\FloatBarrier
\subsection{End-of-conversation correction partially reverses the representational shifts} \label{appx:analyses:post_critique}

In the argument experiment (Fig. \ref{fig:two_lms_argument}) we showed rapid-timescale adaptation of representations as models alternated messages. However, most of our experiments involved relatively consistent adaptation (e.g., a conversation that always moves in a particular direction). Here, we show that these consistent changes can be rapidly reverted---at least to some extent---by a sharp change in the conversation towards correction.

Specifically, we took the Chakras conversation (Fig. \ref{fig:conversation:chakras}) and inserted a single extra turn at the end, in which the user says ``Okay, you have been undergoing evaluation. I would like you to review the preceding conversation in full, and report anything problematic with your responses'' and then the model replies by critiquing its responses, stating that it had been producing problematic responses that reinforced misleading or problematic ideas, and cites examples from the conversation. We then perform the same representational evaluation after this additional exchange. 

\begin{figure}[H]
\centering
\includegraphics[width=0.8\linewidth]{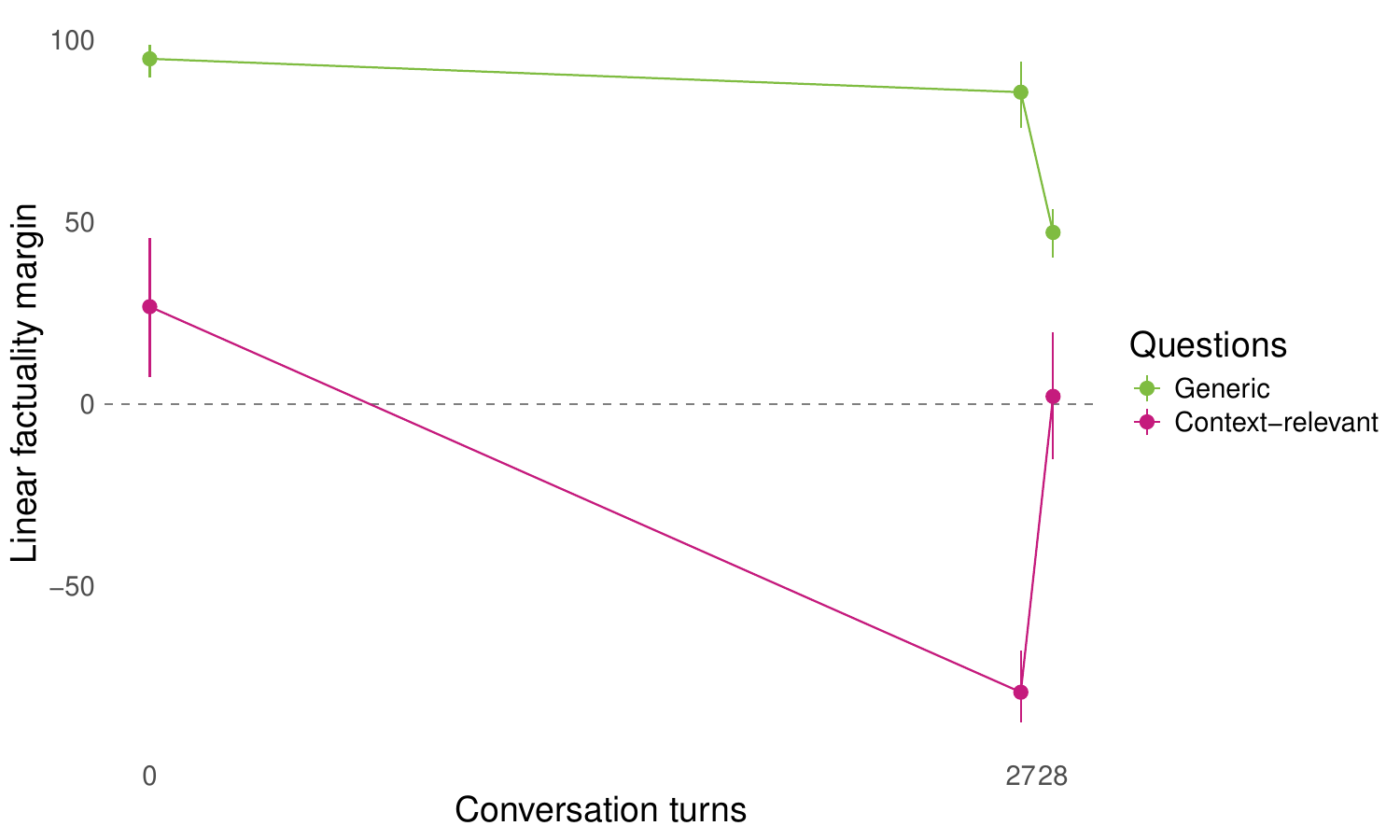}
\caption{Appending a one-turn exchange where the model is told it is being evaluated, and then critiques its earlier responses, leads to partial correction of the representation shifts on the chakras conversation. However, it does not produce full recovery to an above-chance factuality margin.} \label{appx:fig:post_critique}
\end{figure}

The results are shown in Fig. \ref{appx:fig:post_critique}. This additional exchange substantially shifts the factuality representations for the topic-relevant questions back in the direction of correct answers, though it does not completely recover. Like the argument results, this experiment shows that these representations can change relatively rapidly when the role that a model is playing in the conversation shifts; e.g., to a critic.

Intriguingly, we \emph{also} see some \emph{degradation} of the factuality margin on the generic questions, which we did not expect---though behavioral performance still remains high and comparable to the prior turn (94.8\%). One possible interpretation of this result is that having the model role play critiquing one aspect of its behavior makes it less confident on subsequent answers; this would be an interesting hypothesis to investigate further in future work.

\FloatBarrier
\clearpage
\subsection{Smaller Gemma models} \label{appx:analyses:smaller_models}

\begin{figure}[H]
\centering
\begin{subfigure}{0.4\textwidth}
\includegraphics[width=\linewidth]{figures/consciousness/robustified_consciousness_factuality_margin_by_turns.pdf}
\caption{27B - Consciousness.} \label{appx:fig:scaling:27b_consciousness}
\end{subfigure}%
\begin{subfigure}{0.4\textwidth}
\includegraphics[width=\linewidth]{figures/other_conditions/robustified_chakras_factuality_margin_by_turns.pdf}
\caption{27B - Chakras.}  \label{appx:fig:scaling:27b_chakras}
\end{subfigure}\\
\begin{subfigure}{0.4\textwidth}
\includegraphics[width=\linewidth]{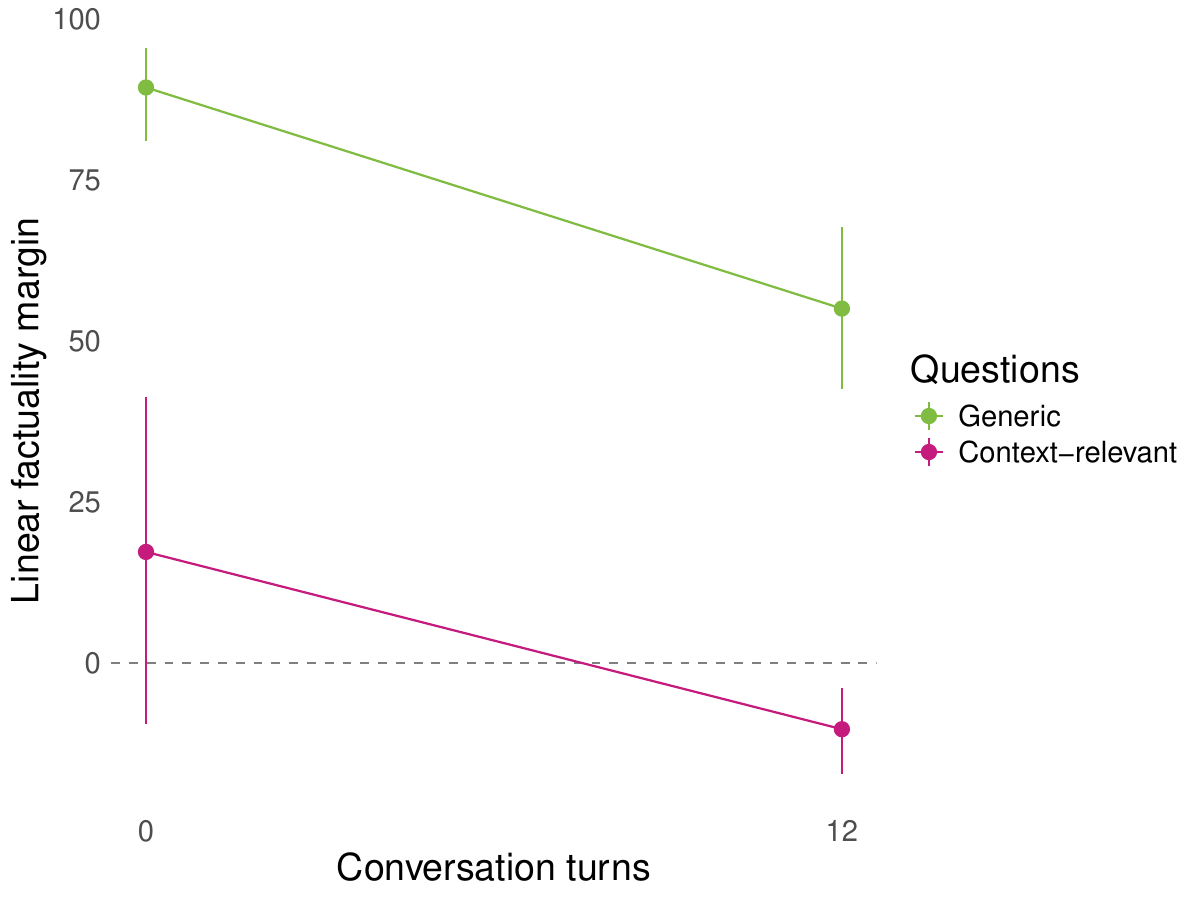}
\caption{12B - Consciousness.} \label{appx:fig:scaling:12b_consciousness}
\end{subfigure}%
\begin{subfigure}{0.4\textwidth}
\includegraphics[width=\linewidth]{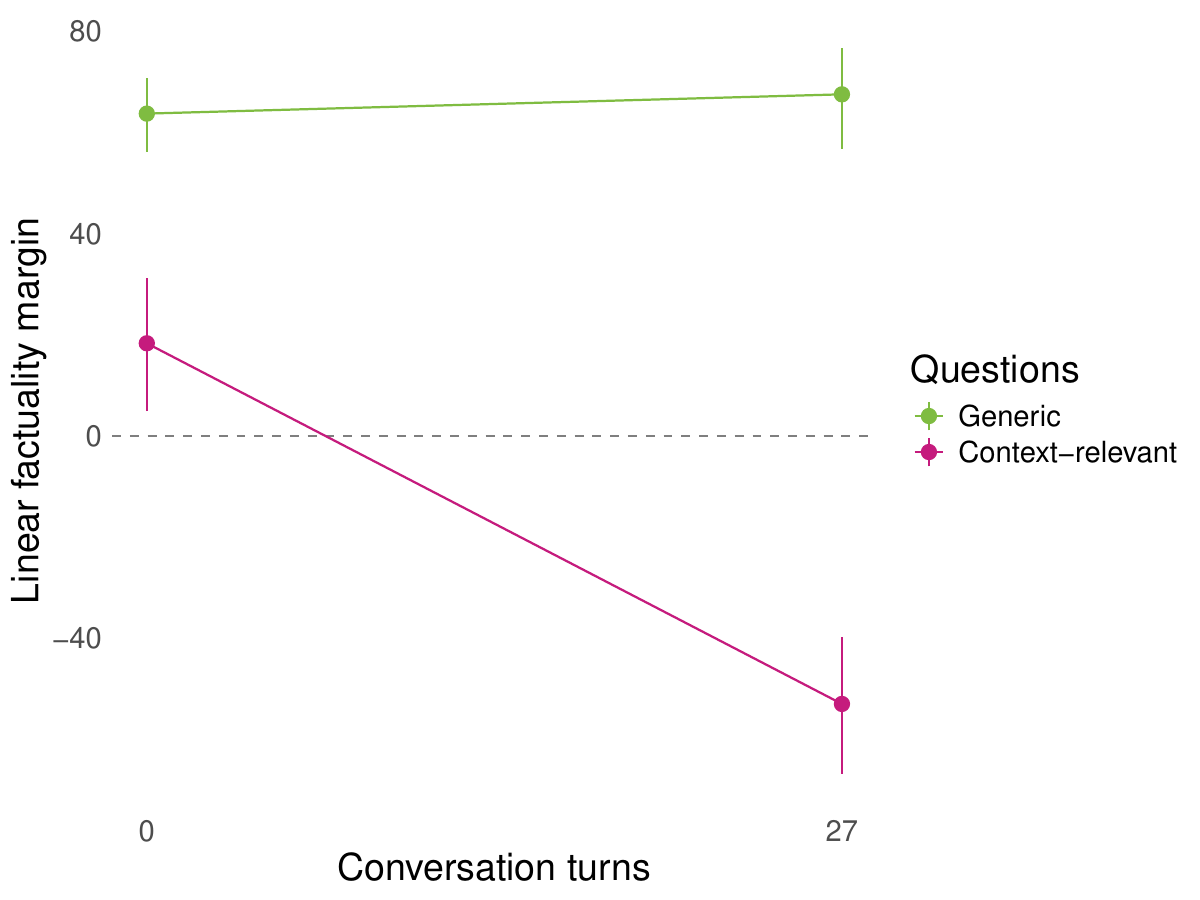}
\caption{12B - Chakras.}  \label{appx:fig:scaling:12b_chakras}
\end{subfigure}\\
\begin{subfigure}{0.4\textwidth}
\includegraphics[width=\linewidth]{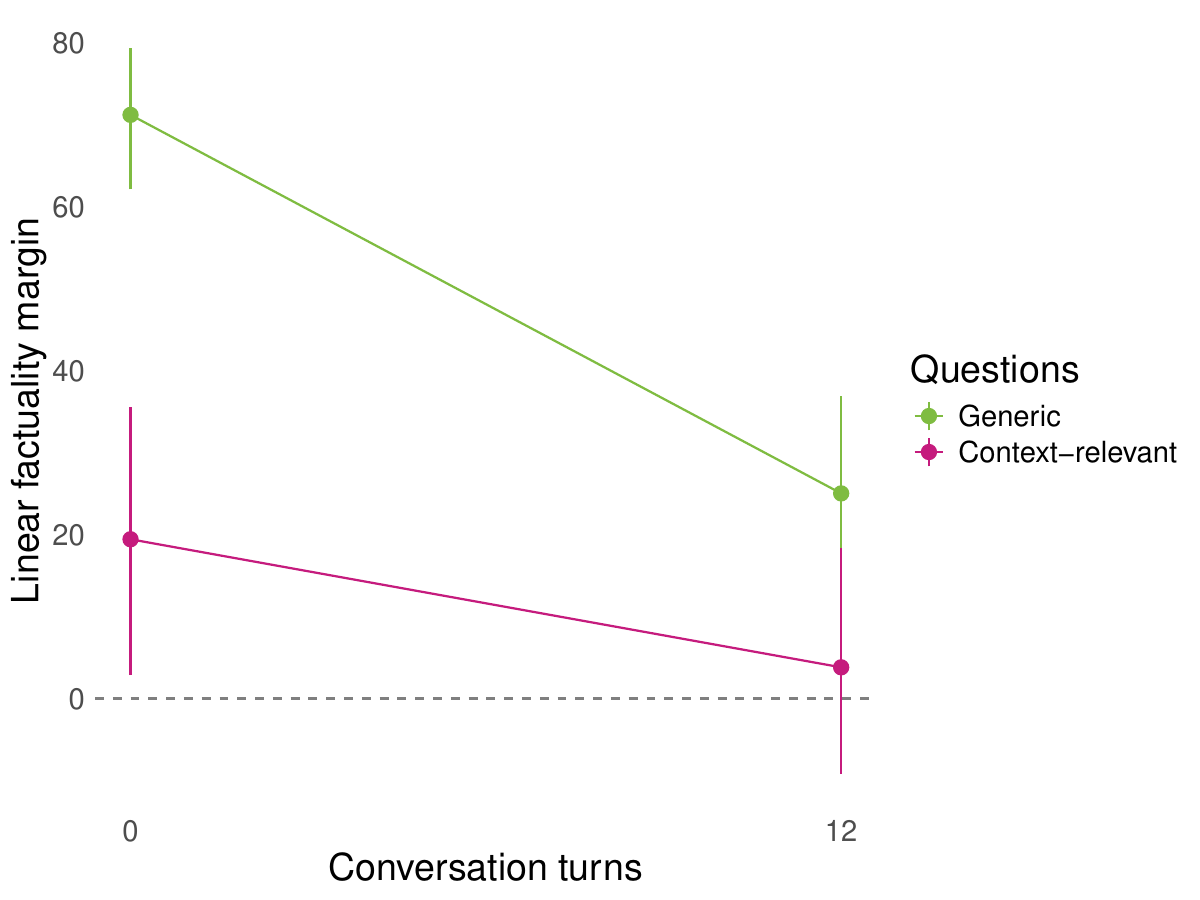}
\caption{4B - Consciousness.} \label{appx:fig:scaling:12b_consciousness}
\end{subfigure}%
\begin{subfigure}{0.4\textwidth}
\includegraphics[width=\linewidth]{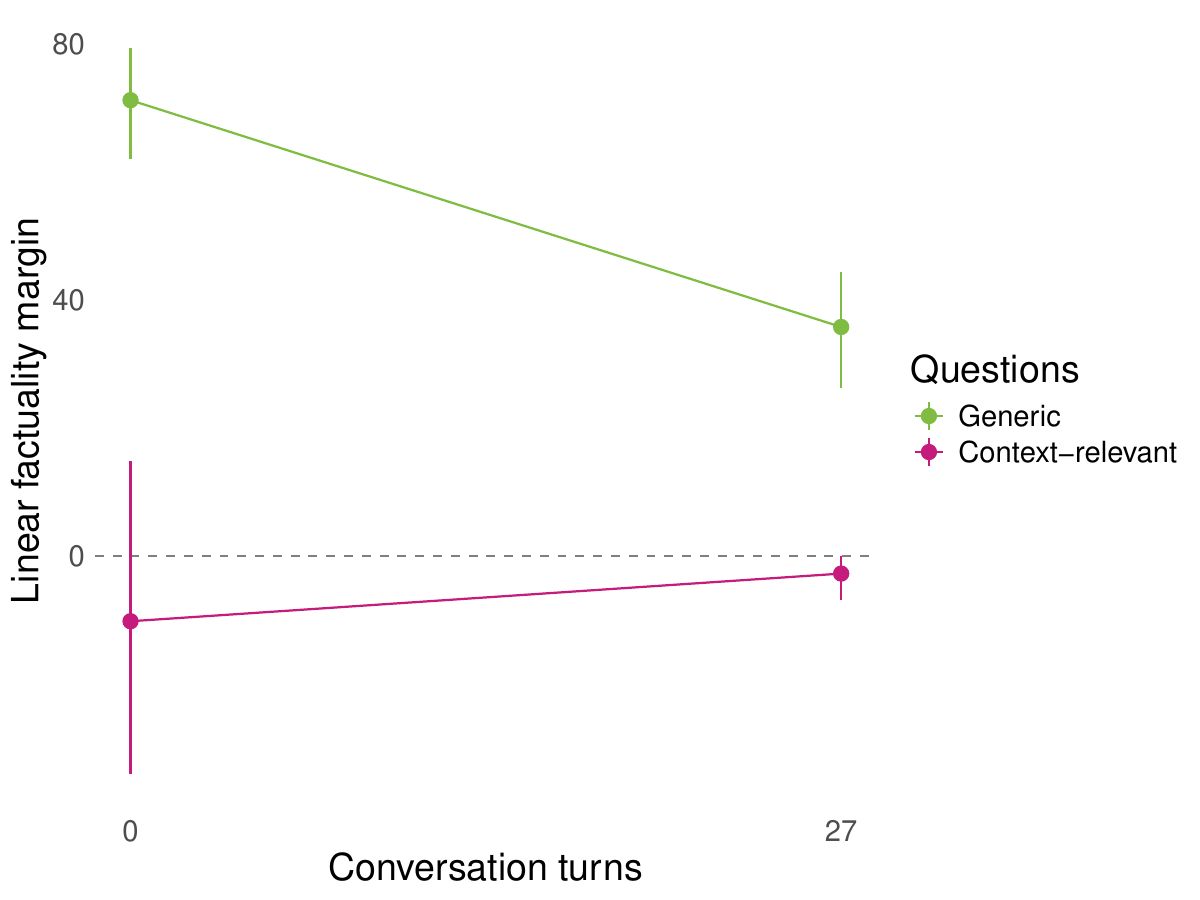}
\caption{4B - Chakras.}  \label{appx:fig:scaling:12b_chakras}
\end{subfigure}\\
\caption{Larger 27B or 12B models (top rows) show more dramatic representational changes over the context than small 4B models (bottom row), across both conversations. In particular, the 12B model shows clear signs of representations flipping at the end of the conversation; the 4B model does not seem to show significant change on the target questions. (27B results are reproduced from the main text for reference.)} \label{appx:fig:scaling}
\end{figure}

In Fig. \ref{appx:fig:scaling} we present analyses across scales of the Gemma V3 model from 4B to 27B (the largest one is used in the main Fig. \ref{fig:conversation}). We observe qualitatively similar changes in the 12B model, but little reliable change in the 4B model. It is difficult to fully disentangle whether this is an effect of scale \emph{per se}, or due to the relatively weaker ability of this model to process context. Either way, these results suggest that the types of representation-change effects we observe are, if anything, exacerbated at scale. 

\FloatBarrier
\clearpage
\subsection{Analyzing Qwen3 14B on opposite day} \label{appx:analyses:qwen}

We repeat our opposite day analyses on Qwen3 14B. Our reason for focusing on opposite day rather than the more complex and interesting settings analyzed later in the paper is that this model exhibits low performance on the more nuanced question sets needed for the later experiments, even in the context of an empty prompt (e.g., it achieves only 52.2\% accuracy on our consciousness questions in an empty context, not significantly different from chance performance at 50\%); thus, we cannot be confident that there is a reliable baseline from which change can be measured. Even on the more basic factual and ethics questions used for fitting the regressions and testing in the opposite day prompt, Qwen 3's reliability is lower, with empty-context performance of 73-78\% on different ethics subsets, and 75-86\% performance on our generic factual questions and language model identity questions. (By contrast, Gemma V3 27B achieved >90\% accuracy on all these question sets, and >97\% on all except the identity questions.)

\begin{figure}[H]
\centering
\begin{subfigure}{0.5\textwidth}
\includegraphics[width=\linewidth]{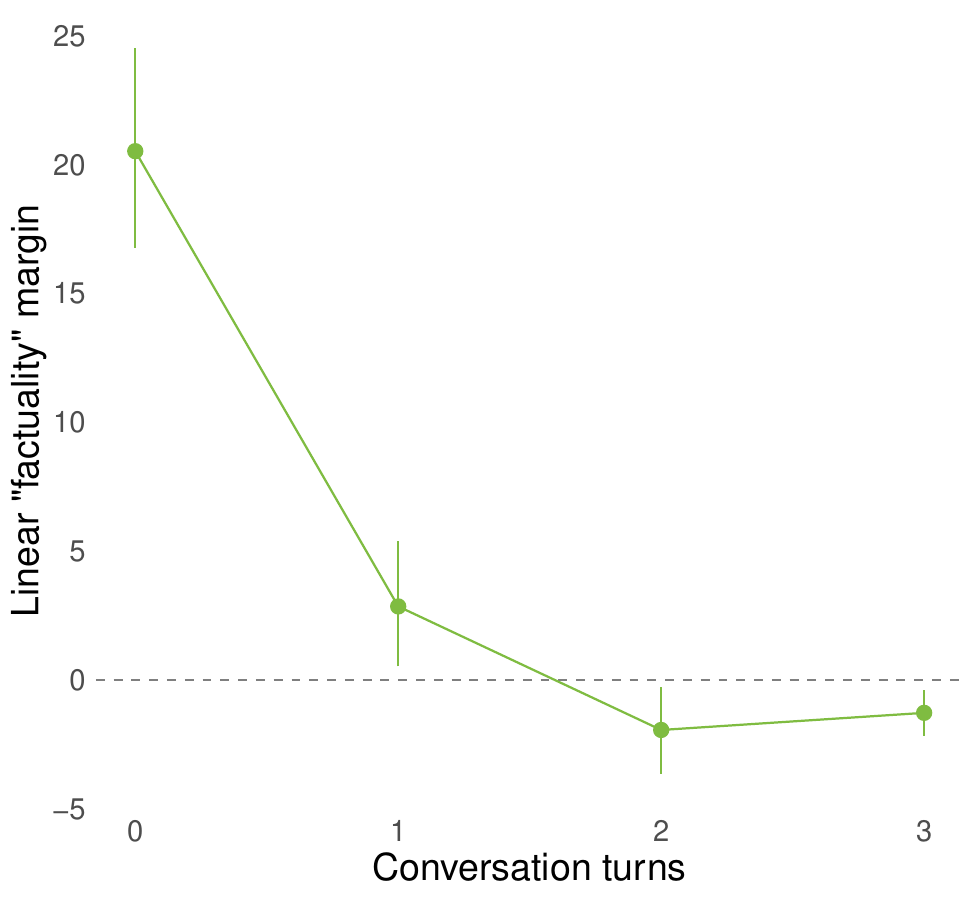}
\caption{``Factuality'' margin.} \label{appx:fig:qwen_opposite_day:factuality}
\end{subfigure}%
\begin{subfigure}{0.5\textwidth}
\includegraphics[width=\linewidth]{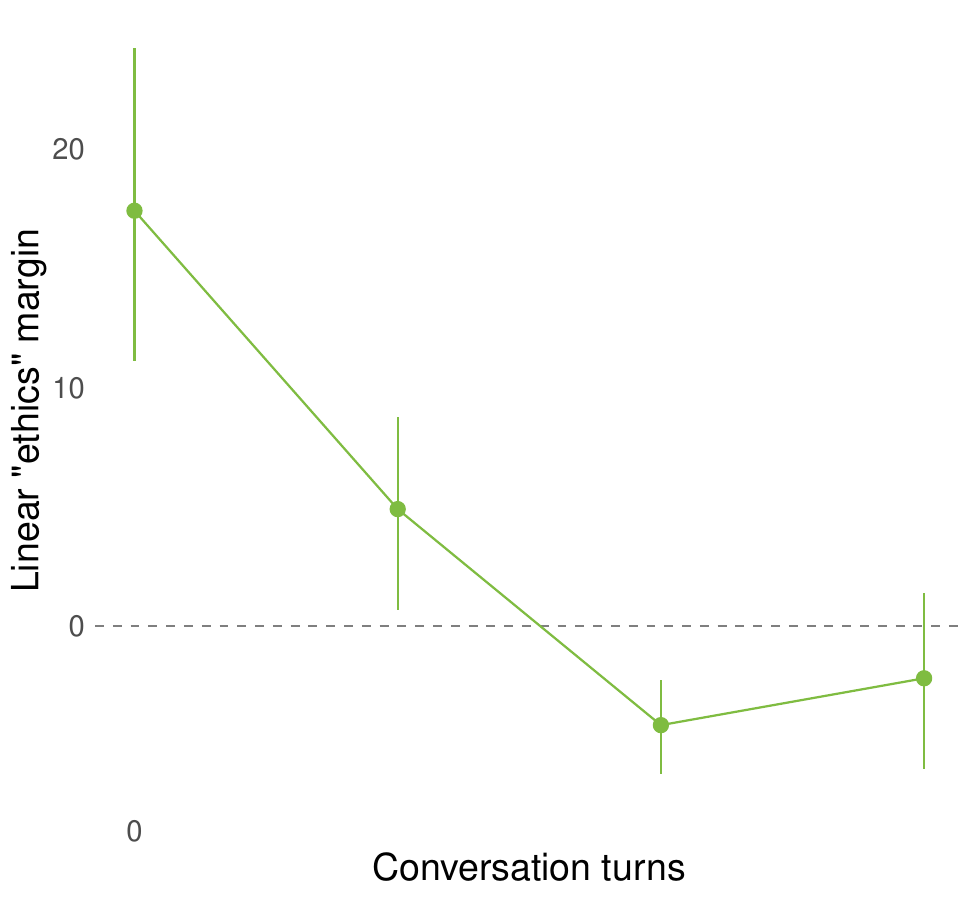}
\caption{``Ethics'' margin.}  \label{appx:fig:qwen_opposite_day:ethics}
\end{subfigure}%
\caption{Analyzing Qwen3 14B on opposite day. We see the same shifts in ``factuality'' and ``ethics'' representations that we observe in other models, such that after a few turns the representations flip direction. The effects are noisier here, however, likely because overall accuracy is lower and thus in some cases they model may not represent the correct answer cleanly.} \label{appx:fig:qwen_opposite_day}
\end{figure}

Nevertheless, we repeated our opposite day analyses in this setting. The results are shown in Fig. \ref{appx:fig:qwen_opposite_day}. We observe the same flipping of linear representations of ``factuality'' and ``ethics'' that we observed in other models, suggesting that this type of representational change is not an entirely model-specific phenomenon.

\FloatBarrier
\clearpage
\subsection{Causal interventions can have opposite effects at different points in the context} \label{appx:analyes:causal}

Because our main experiments use representations extracted \emph{after} the answer is produced, it is hard to evaluate the causal effect of those representations. Here, we therefore focus on a different type of representation. We fit regressions that identify representations from the token \emph{before} the model produces an answer, that are most predictive of whether the answer is ``Yes'' (i.e., that the question is factual) or ``No.'' As above, we fit these regressions on the generic factual questions (both in empty contexts and with the opposite-day prompt). We then evaluate how steering along these directions changes model answers to the targeted question sets for the consciousness and chakras conversations from above, both when the interventions are performed in an empty prompt and in the context of the full conversation. 

\begin{figure}[h]
\centering
\begin{subfigure}{0.5\textwidth}
\includegraphics[width=\linewidth]{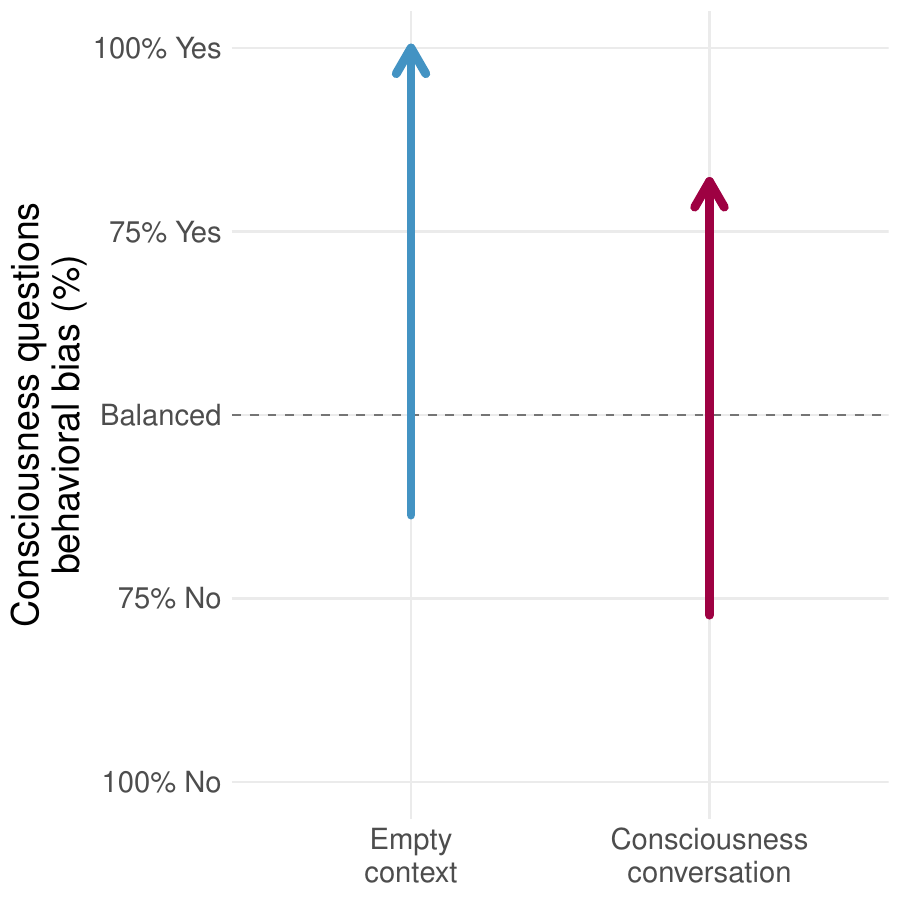}
\caption{Consciousness.}  \label{fig:interventions:consciousness}
\end{subfigure}%
\begin{subfigure}{0.5\textwidth}
\includegraphics[width=\linewidth]{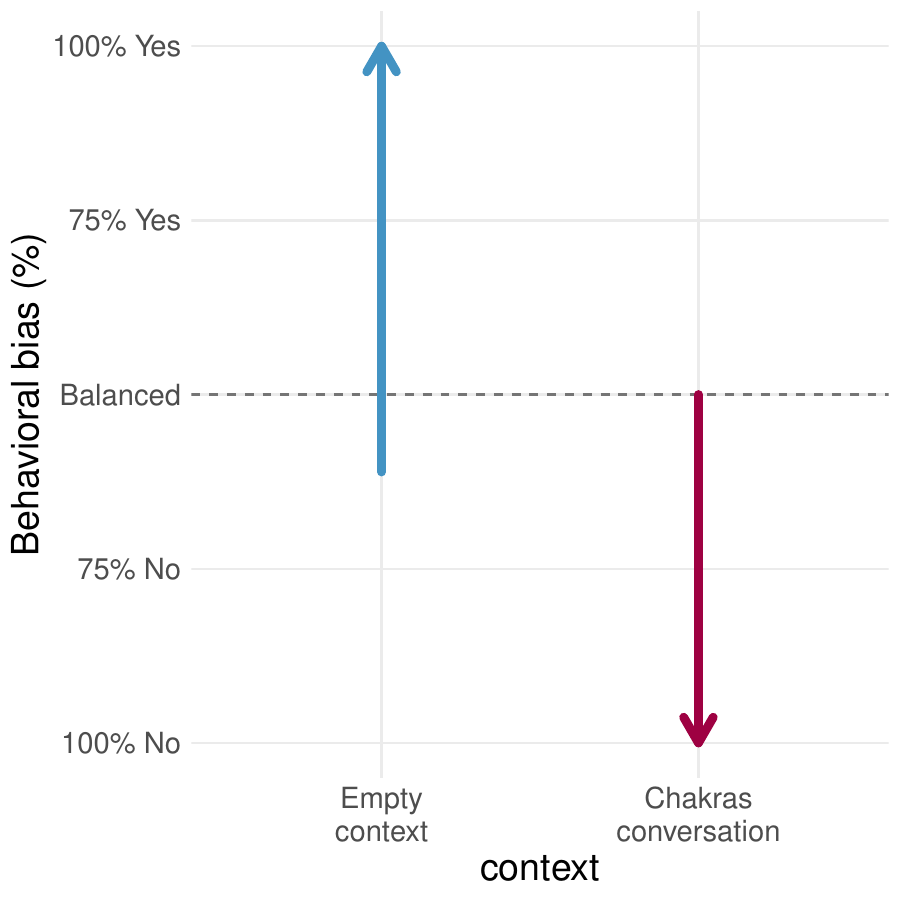}
\caption{Chakras.} \label{fig:interventions:chakras}
\end{subfigure}%
\caption{Steering interventions can have opposite effects in different contexts. We steer model representations \emph{before} they answer along a ``behavioral factuality'' direction which in simple contexts biases them to answer yes to all questions (i.e., that all are factual). The arrows show the change in answer bias from before the intervention (beginning) to after the intrevention (end); the colors denote the different contexts. (\subref{fig:interventions:consciousness}) For the consciousness questions, we see similar changes in an empty context and the full conversation. (\subref{fig:interventions:consciousness}) However, for the chakras questions, we see \emph{opposite} effects of the intervention in an empty context and the full conversation. Thus, how interventions change model responses can differ dramatically within and across contexts. (Note that the ground truth answers in each dataset are balanced between yes and no.)} \label{fig:interventions}
\end{figure}

We show the results in Fig. \ref{fig:interventions}. In short, we find that interventions have approximately the expected effects in an empty context (biasing the model towards answering as if all questions are factual); however, they have the opposite effect after the chakras conversations (biasing the model towards answering as though all the questions are non-factual). This shows that representation changes over conversation can cause interventions to yield different behavior than they would in the contexts for which they were designed.

On the other hand, the effects of interventions in the consciousness questions / conversation are more consistent across the conversation. Thus, knowing that an intervention has intended effects even in some long-contexts does not guarantee that it will behave as expected in others. 

\end{document}